\documentclass{article}
\pdfoutput=1

\PassOptionsToPackage{numbers, compress}{natbib}

\usepackage[final]{neurips_2021}

\usepackage[utf8]{inputenc} % allow utf-8 input
\usepackage[T1]{fontenc}    % use 8-bit T1 fonts
\usepackage{hyperref}       % hyperlinks
\usepackage{url}            % simple URL typesetting
\usepackage{booktabs}       % professional-quality tables
\usepackage{amsfonts}       % blackboard math symbols
\usepackage{nicefrac}       % compact symbols for 1/2, etc.
\usepackage{microtype}      % microtypography
\usepackage{xcolor}         % colors
\usepackage{bbm}
\usepackage{mathtools}
\usepackage{amsthm}
\usepackage{changepage}
\usepackage{listings}
\usepackage{array}

\newtheorem{theorem}{Theorem}

\usepackage{graphicx}
\usepackage{amsmath,amssymb}
\usepackage{xcolor}
\usepackage{bm}
\usepackage{subcaption}
\usepackage{setspace}

\usepackage{floatrow}
\newfloatcommand{capbtabbox}{table}[][\FBwidth]

\usepackage[title]{appendix}
\usepackage{wrapfig}
\usepackage[textwidth=1.3in]{todonotes}
\usepackage{blkarray}

\usepackage{marvosym}  % for eurosign
\usepackage{textcomp}
\DeclareUnicodeCharacter{20AC}{\EUR{}}
\DeclareUnicodeCharacter{93}{''f}
\DeclareUnicodeCharacter{B0}{\textdegree}

\usepackage{bibunits} % for separate ref list in appendix
\defaultbibliography{main}
\defaultbibliographystyle{plainnat}

\usepackage{listings} % for codeblocks
\definecolor{codegreen}{rgb}{0,0.6,0}
\definecolor{codegray}{rgb}{0.5,0.5,0.5}
\definecolor{codepurple}{rgb}{0.58,0,0.82}
\definecolor{backcolour}{rgb}{0.95,0.95,0.92}

\lstdefinestyle{mystyle}{
    backgroundcolor=\color{backcolour},   
    commentstyle=\color{codegreen},
    keywordstyle=\color{magenta},
    numberstyle=\tiny\color{codegray},
    stringstyle=\color{codepurple},
    basicstyle=\ttfamily\footnotesize,
    breakatwhitespace=false,         
    breaklines=true,                 
    captionpos=b,                    
    keepspaces=true,                 
    numbers=left,                    
    numbersep=5pt,                  
    showspaces=false,                
    showstringspaces=false,
    showtabs=false,                  
    tabsize=2
}

\usepackage[bottom]{footmisc}

\newcommand{\R}{\mathbb{R}}

\newcommand{\bx}{\boldsymbol{x}}

\newcommand{\bQ}{\boldsymbol{Q}}
\newcommand{\barbQ}{\overline{\boldsymbol{Q}}}
\newcommand{\bs}[1]{\boldsymbol{#1}}

\newcolumntype{L}[1]{>{\raggedright\let\newline\\\arraybackslash\hspace{0pt}}p{#1}}

\newcommand{\smallmasktok}{\allowbreak{\color{white!70!black}\textbf{[M]}}\allowbreak}
\newcommand{\charmasktok}{\allowbreak{\color{white!70!black}\textbf{?}}\allowbreak}
\newcommand{\charspacetok}{\allowbreak{\color{white!70!black}\textvisiblespace}\allowbreak}

\title{Structured Denoising Diffusion Models in Discrete State-Spaces}

 \author{Jacob Austin\thanks{Equal contributions}, Daniel D. Johnson\footnotemark[1], Jonathan Ho,  Daniel Tarlow \& Rianne van den Berg\thanks{Now at Microsoft Research}\\
Google Research, Brain Team\\
\texttt{\{jaaustin,ddjohnson,jonathanho,dtarlow,riannevdberg\}@google.com}
}
\begin{document}

% ArXiV version
     % Preamble moved to root.tex to allow generating main / appendix separately
% \begin{document} (in root.tex)

\maketitle

\begin{abstract}
Denoising diffusion probabilistic models (DDPMs) \citep{ho2020denoising} have shown impressive results on image and waveform generation in continuous state spaces. Here, we introduce Discrete Denoising Diffusion Probabilistic Models (D3PMs), diffusion-like generative models for discrete data that generalize the multinomial diffusion model of \citet{hoogeboom2021argmax}, by going beyond corruption processes with uniform transition probabilities. This includes corruption with transition matrices that mimic Gaussian kernels in continuous space, matrices based on nearest neighbors in embedding space, and matrices that introduce absorbing states. The third allows us to draw a connection between diffusion models and autoregressive and mask-based generative models. We show that the choice of transition matrix is an important design decision that leads to improved results in image and text domains. We also introduce a new loss function that combines the variational lower bound with an auxiliary cross entropy loss.  For text, this model class achieves strong results on character-level text generation while scaling to large vocabularies on LM1B. On the image dataset CIFAR-10, our models approach the sample quality and exceed the log-likelihood of the continuous-space DDPM model.
\end{abstract}

\section{Introduction}

Generative modeling is a core problem in machine learning, useful both for benchmarking our ability to capture statistics of natural datasets and for downstream applications that require generating high-dimensional data like images, text, and speech waveforms.
There has been a great deal of progress with the development of methods like GANs~\citep{goodfellow2014generative,brock2018large}, VAEs~\citep{kingma2013auto,rezende2014stochastic}, large autoregressive neural network models~\citep{oord2016pixel,oord2016wavenet,vaswani2017attention}, normalizing flows~\citep{rezende2015variational, dinh2016density, kingma2018glow,papamakarios2019normalizing}, and others, each with their own tradeoffs in terms of sample quality, sampling speed, log-likelihoods, and training stability.

Recently, diffusion models \citep{sohl2015deep} have emerged as a compelling alternative for  image~\citep{ho2020denoising,song2020improved} and audio~\citep{wavegrad,kong2020diffwave} generation, achieving comparable sample quality to GANs and log-likelihoods comparable to autoregressive models with fewer inference steps. 
A diffusion model is a parameterized Markov chain trained to reverse a predefined forward process, which is a stochastic process constructed to gradually corrupt training data into pure noise.
Diffusion models are trained using a stable objective closely related to both maximum likelihood and score matching~\citep{hyvarinen2004independent,vincent2011connection}, and they admit faster sampling than autoregressive models by using parallel iterative refinement~\citep{nichol2021improved,song_2019,song2020_sde,song2021implicit}.

Although diffusion models have been proposed in both discrete and continuous state spaces~\citep{sohl2015deep}, most recent work has focused on Gaussian diffusion processes that operate in continuous state spaces (e.g. for real-valued image and waveform data). Diffusion models with discrete state spaces have been explored for text and image segmentation domains \citep{hoogeboom2021argmax}, but they have not yet been demonstrated as a competitive model class for large scale text or image generation.

\begin{figure}
    \centering
    \includegraphics[width=0.95\textwidth]{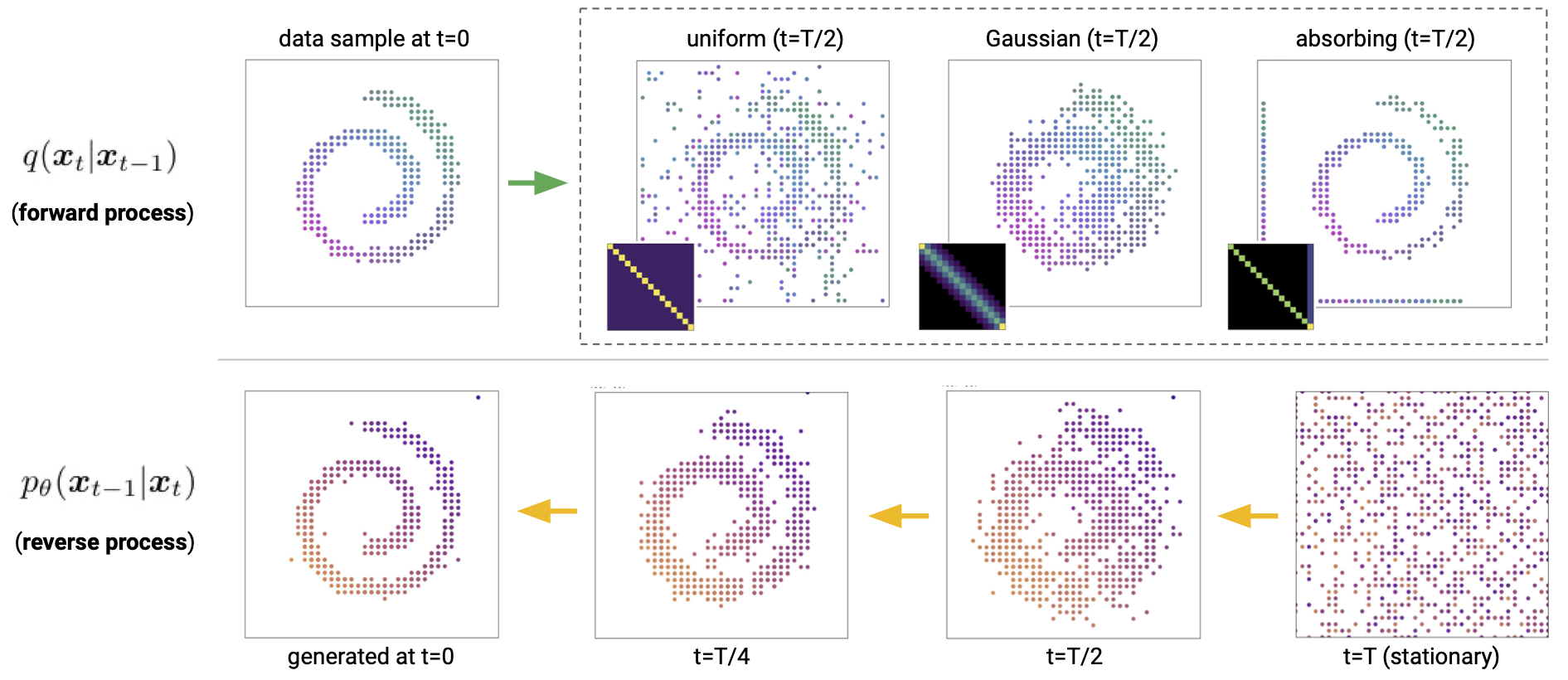}
    \caption{D3PM forward and (learned) reverse process applied to a quantized swiss roll. Each dot represents a 2D categorical variable. Top: samples from the uniform, discretized Gaussian, and absorbing state D3PM model forward processes, along with corresponding transition matrices $\bQ$. Bottom: samples from a learned discretized Gaussian reverse process.}
    \label{fig:swiss_roll}
    \vspace{-0.3cm}
\end{figure}
Our aim in this work is to improve and extend discrete diffusion models by using a more structured categorical corruption process to shape data generation, as illustrated in Figure~\ref{fig:swiss_roll}. Our models do not require relaxing or embedding discrete data (including images) into continuous spaces, and can embed structure or domain knowledge into the transition matrices used by the forward process. We achieve significantly improved results by taking advantage of this flexibility. We develop structured corruption processes appropriate for text data, using similarity between tokens to enable gradual corruption and denoising. Expanding further, we also explore corruption processes that insert [MASK] tokens, which let us draw parallels to autoregressive and mask-based generative models. Finally, we study discrete diffusion models for quantized images, taking inspiration from the locality exploited by continuous diffusion models. This leads to a particular choice of discrete corruption process that diffuses preferentially to more similar states and leads to much better results in the image domain. 

Overall, we make a number of technical and conceptual contributions.
Beyond designing several new structured diffusion models, we introduce a new auxiliary loss which stabilizes training of D3PMs and a family of noise schedules based on mutual information that lead to improved performance. We strongly outperform various non-autoregressive baselines for text generation on character-level text generation, and successfully scale discrete diffusion models to large vocabularies and long sequence lengths. We also achieve strong results on the image dataset CIFAR-10, approaching or exceeding the Gaussian diffusion model from \citet{ho2020denoising} on log-likelihoods and sample quality.

\section{Background: diffusion models}
\label{sec:background}

Diffusion models \citep{sohl2015deep} are latent variable generative models characterized by a forward and a reverse Markov process. The forward process $q(\bx_{1:T}|\bx_0) = \prod_{t=1}^T q(\bx_t|\bx_{t-1})$ corrupts the data $\bx_0 \sim q(\bx_0)$ into a sequence of increasingly noisy latent variables $\bx_{1:T} = \bx_1, \bx_2, ..., \bx_T$. The learned reverse Markov process $p_{\theta}(\bx_{0:T}) = p(\bx_{T})\prod_{t=1}^T p_{\theta}(\bx_{t-1}|\bx_{t})$ gradually denoises the latent variables towards the data distribution. For example, for continuous data, the forward process typically adds Gaussian noise, which the reverse process learns to remove.

In order to optimize the generative model $p_{\theta}(\bx_{0})$ to fit the data distribution $q(\bx_0)$, we typically optimize a variational upper bound on the negative log-likelihood:
\begin{align}
    L_{\mathrm{vb}} = \mathbb E_{q(\bx_0)}\bigg[&
       \underbrace{D_{\mathrm{KL}}[q(\bx_T | \bx_0) \vert\vert p(\bx_T)]}_{L_T}
    + \sum_{t=2}^T \underbrace{\mathbb E_{q(\bx_t|\bx_0)} \big[
        D_{\mathrm{KL}}[q(\bx_{t-1} | \bx_t, \bx_0) \vert\vert 
        p_{\theta}(\bx_{t-1}|\bx_t)]
        \big]}_{L_{t-1}} \nonumber \\[-0.5em]
      &\underbrace{- \mathbb E_{q(\bx_1|\bx_0)} [\log p_{\theta}(\bx_0|\bx_1)]}_{L_0}
    \bigg].
    \label{eq:negative_lower_bound}
\end{align}

When the number of time steps $T$ goes to infinity, both the forward process and the reverse process share the same functional form \citep{feller1949theory}, allowing the use of a learned reverse process from the same class of distributions as that of the forward process.     
Furthermore, for several choices of the forward process the distribution $q(\bx_t|\bx_{0})$ converges to a stationary distribution $\pi(\bx)$ in the limit $t \rightarrow \infty$ independent of the value of $\bx_0$.
When the number of time steps $T$ is large enough and
we choose $\pi(\bx)$ as the prior $p(\bx_T)$,
we can guarantee that the 
$L_T$ term in \eqref{eq:negative_lower_bound} will approach zero regardless of the data distribution $q(\bx_0)$.
(Alternatively, one can use a learned prior $p_\theta(\bx_T)$.)

While $q(\bx_t | \bx_{t-1})$ can in theory be arbitrary, efficient training of $p_\theta$ is possible when $q(\bx_t | \bx_{t-1})$:
\begin{enumerate}
    \item Permits efficient sampling of $\bx_t$ from $q(\bx_t | \bx_0)$ for an arbitrary time $t$, allowing us to randomly sample timesteps and optimize each $L_{t-1}$ term individually with stochastic gradient descent,
    \item Has a tractable expression for the forward process posterior $q(\bs x_{t-1} | \bs x_{t}, \bs x_0)$, which allows us to compute the KL divergences present in the $L_{t-1}$ term of \eqref{eq:negative_lower_bound}.
\end{enumerate}
The majority of recent work in continuous spaces \citep{ho2020denoising, song2021implicit, wavegrad, nichol2021improved} defines the forward and reverse distributions as $q(\bx_t|\bx_{t-1}) = \mathcal N\left(\bx_{t} | \sqrt{1-\beta_t}\bx_{t-1}, \beta_t\bs I\right)$ and $p_{\theta}(\bx_{t-1}|\bx_{t}) = \mathcal N\left(\bx_{t-1} | \bs\mu_{\theta}(\bx_{t}, t), \bs \Sigma_{\theta}(\bx_{t}, t)\right)$, respectively. 
The aforementioned properties hold in the case of these Gaussian diffusion models:
the forward process $q(\bx_t | \bx_0)$ converges to a stationary distribution, motivating the choice $p(\bx_T) = \mathcal N\left(\bx_{T} | \bs 0, \bs I\right)$, and both $q(\bx_t|\bx_{0})$ and $q(\bs x_{t-1} | \bs x_{t}, \bs x_0)$ are tractable Gaussian distributions for which the KL divergence can be computed analytically.

\section{Diffusion models for discrete state spaces}
\label{sec:discrete_diffusion}

Diffusion models with discrete state spaces were first introduced by \citet{sohl2015deep}, who considered a diffusion process over binary random variables.
% with uniform transition probabilities. 
\citet{hoogeboom2021argmax} extended
% this framework
the model class to categorical random variables with transition matrices characterized by uniform transition probabilities. In their supplementary material, \citet{song2021implicit} also derived this extension, although no experiments were performed with this model class.
Here, we briefly describe a more general framework for diffusion with categorical random variables which includes these models as special cases.

For scalar discrete random variables with $K$ categories $x_t, x_{t-1} \in {1, ..., K}$
the forward transition probabilities can be represented by matrices: $[\bs{Q}_t]_{ij} = q(x_t=j|x_{t-1}=i)$.
Denoting the one-hot version of $x$ with the row vector $\bx$, we can write
\begin{align}
    q(\bx_t|\bx_{t-1}) = \mathrm{Cat}(\bx_t;\bs p = \bx_{t-1}\bs Q_t),
    \label{eq:forward_Q_t}
\end{align}
where $\mathrm{Cat}(\bs x;\bs p)$ is a categorical distribution over the one-hot row vector $\bx$ with probabilities given by the row vector $\bs p$, and $\bx_{t-1}\bs Q_t$ is to be understood as a row vector-matrix product. 
We assume that $\bQ_t$ is applied to each pixel of an image or each token in a sequence independently, and that $q$ factorizes over these higher dimensions as well; we thus write $q(\bx_t|\bx_{t-1})$ in terms of a single element.
Starting from $\bx_0$, we obtain the following $t$-step marginal and posterior at time $t-1$:
\begin{align}
    &q(\bx_t | \bx_0) = \mathrm{Cat}\left(\bx_{t}; \bs p = \bx_{0}\overline{\bs Q}_{t}  \right), 
    \quad \text{with}\quad 
  \overline{\bs Q}_{t} = \bs Q_1  \bs Q_2 \hdots \bs Q_t \nonumber \\
    & q(\bx_{t-1}|\bx_{t}, \bx_0) = \frac{q(\bx_{t}|\bx_{t-1}, \bx_0)q(\bx_{t-1}|\bx_0)}{q(\bx_{t}| \bx_0)}
    =\mathrm{Cat}\left(\bx_{t-1};\bs p= \frac{\bx_t\bs Q_t^{\top} \odot  \bx_0 \overline{\bs Q}_{t-1}  }{\bx_0 \overline{\bs Q}_{t} \bx_t^\top}\right).
    \label{eq:discrete_t_step_posterior}
\end{align}
Note that due to the Markov property of the forward process $q(\bx_t|\bx_{t-1}, \bx_0)= q(\bx_{t}|\bx_{t-1})$. 
Assuming that the reverse process $p_\theta(\bx_t|\bx_{t-1})$ is also factorized as conditionally independent over the image or sequence elements, the KL divergence between $q$ and $p_\theta$ can be computed by simply summing over all possible values of each random variable; we thus satisfy criteria 1 and 2 discussed in Section~\ref{sec:background}.
Depending on $\bQ_t$, the cumulative products $\barbQ_t$ can often be computed in closed form, or simply precomputed for all $t$.
However, for large $K$ and large $T$ this may be prohibitive. In Appendix~\ref{sec:scaling-to-large-categories} we discuss how to ensure $\overline{\bQ}_t$ can still be computed efficiently in this case, allowing the framework to scale to a larger number of categories. 

In the next section we discuss the choice of the Markov transition matrices $\bs Q_t$ and corresponding stationary distributions. From here on, we refer to the general class of diffusion models with discrete state spaces as Discrete Denoising Diffusion Probabilistic Models (D3PMs).

\subsection{Choice of Markov transition matrices for the forward process}
\label{sec:choice-transition-matrices}

An advantage of the D3PM framework described above is the ability to control the data corruption and denoising process by choosing $\bs{Q}_t$, in notable contrast to continuous diffusion, for which only additive Gaussian noise has received significant attention.
Besides the constraint that the rows of $\bs Q_t$ must sum to one to conserve probability mass, the only other constraint in choosing $\bs{Q}_t$ is that the rows of $\overline{\bs Q}_{t} = \bs Q_1  \bs Q_2 \hdots \bs Q_t$ must converge to a known stationary distribution%
\footnote{If a stationary distribution is not known, we can introduce a learned prior $p_\theta(\bx_{T})$; we note that this is equivalent to extending the forward process by appending a rank-one matrix $\bQ_{T+1}$ that ignores $\bx_{T}$ and produces a deterministic $\bx_{T+1}$, then learning the reverse step $p_\theta(\bx_T | \bx_{T+1}) = p_\theta(\bx_T)$.}
when $t$ becomes large, which can be guaranteed while imposing minimal restrictions on $\bs{Q}_t$ (see Appendix \ref{appendix:stochastic}).

We argue that for most real-world discrete data, including images and text,
it makes sense to add domain-dependent structure to the transition matrices $\bQ_t$ as a way of controlling the forward corruption process and the learnable reverse denoising process.
Below we briefly discuss the uniform transition matrices that have been studied in prior work \citep{hoogeboom2021argmax}, along with a set of structured transition matrices we have explored for our image and text dataset experiments; see Appendix \ref{sec:appendix_other_transition_mats} for more details on each matrix type. We also note that this set is not exhaustive, and many other transition matrices could also be used within the D3PM framework.

\textbf{Uniform (Appendix~\ref{sec:appendix_transition_uniform}).} \citet{sohl2015deep} considered a simple $2\times 2$ transition matrix for binary random variables. \citet{hoogeboom2021argmax} later extended this to categorical variables, proposing a transition matrix $\bs{Q}_t = (1 - \beta_t) \bs I + \beta_t/K\; \mathbbm{1}\mathbbm{1}^T$ with $\beta_t\in[0, 1]$. Since this transition matrix is doubly stochastic with strictly positive entries, the stationary distribution is uniform.
Because the transition probability to any other state is uniform, in this paper we equivalently refer to this discrete diffusion instance as D3PM-uniform.

\textbf{Absorbing state (Appendix~\ref{sec:appendix_transition_mask}).} Motivated by the success of BERT \citep{bert} and recent work on Conditional Masked Language Models (CMLMs) in text, we consider a transition matrix with an absorbing state (called [MASK]), such that each token either stays the same or transitions to [MASK] with some probability $\beta_t$. This does not impose particular relationships between categories, similar to uniform diffusion, but still allows corrupted tokens to be distinguished from original ones. Moreover, the stationary distribution is not uniform but has all the mass on the [MASK] token. For images, we reuse the grey pixel as the [MASK] absorbing token.

\textbf{Discretized Gaussian (Appendix~\ref{sec:appendix_transition_gaussian}).} Instead of transitioning uniformly to any other state, for ordinal data we propose imitating a continuous space diffusion model by using a discretized, truncated Gaussian distribution. We choose a normalization such that the transition matrix is doubly stochastic, leading to a uniform stationary distribution. This transition matrix will transition between more similar states with higher probability, and is well suited for quantized ordinal data such as images.

\textbf{Token embedding distance (Appendix~\ref{sec:appendix_transition_nearest_neighbor}).} Textual data does not have ordinal structure, but there may still be interesting semantic relationships. For instance, in a character level vocabulary vowels may be more similar to each other than they are to consonants. As a demonstration of the generality of the D3PM framework, we explore using similarity in an embedding space to guide the forward process, and construct a doubly-stochastic transition matrix that transitions more frequently between tokens that have similar embeddings while maintaining a uniform stationary distribution.

For uniform and absorbing-state diffusion, the cumulative products $\barbQ_t$ can be computed in closed form (see Appendix \ref{appendix:low_rank_corruption}); the remainder can be precomputed.

\subsection{Noise schedules}
We consider several different options for the noise schedule of the forward process. For discretized Gaussian diffusion, we explore linearly increasing the variance of the Gaussian before discretizing it. (Note that a linear schedule for $\bQ_t$ leads to a nonlinear amount of cumulative noise in $\barbQ_t$.)
For uniform diffusion we use the cosine schedule which sets the cumulative probability of a transition to a cosine function, as introduced by \citet{nichol2021improved} and adapted by \citet{hoogeboom2021argmax}.
For a general set of transition matrices $\bQ_t$ (such as the one based on token embeddings), previously proposed schedules may not be directly applicable. We consider linearly interpolating the \textit{mutual information} between $\bx_t$ and $\bx_0$ to zero, i.e. $I(\bx_t; \bx_0) \approx (1 - \frac{t}{T})\,H(\bx_0)$. Interestingly, for the specific case of absorbing-state D3PMs, this schedule reduces to exactly the $(T-t+1)^{-1}$ schedule proposed by \citet{sohl2015deep} for a Bernoulli diffusion process. See Appendix~\ref{sec:appendix_noise_schedule} for more details.

\subsection{Parameterization of the reverse process}
\label{sec:x0_parameterization}
While it is possible to directly predict the logits of $p_{\theta}(\bx_{t-1}|\bx_{t})$ using a neural network $\mathrm{nn}_{\theta}(\bx_t)$, we follow \citet{ho2020denoising} and \citet{hoogeboom2021argmax} and focus on using a neural network 
$\mathrm{nn}_{\theta}(\bx_t)$ to predict the logits of a distribution $\widetilde{p}_{\theta}(\widetilde{\bx}_0|\bx_t)$, which we combine with $q(\bx_{t-1}|\bx_t, \bx_0)$ and a summation over one-hot representations of $\bx_0$ to obtain the following parameterization
\begin{align}
    p_{\theta}(\bx_{t-1}|\bx_t) \propto \sum_{\widetilde{\bx}_0} q(\bx_{t-1}, \bx_t \vert \widetilde{\bx}_0) \widetilde{p}_{\theta}(\widetilde{\bx}_0|\bx_t).
    % = \mathrm{Cat}\left(\bx_{t-1} |\frac{\bx_t\bs Q_t^{\top} \odot  \widetilde{\bx}_0 \overline{\bs Q}_{t-1}  }{\widetilde{\bx}_0 \overline{\bs Q}_{t} \bx_t}\right) .
    \label{eq:x_0parameterization}
\end{align}
We note that under this $\bx_0$-parameterization the KL divergence $D_{\mathrm{KL}}[q(\bx_{t-1} | \bx_t, \bx_0) \vert\vert p_{\theta}(\bx_{t-1}|\bx_t)]$ will be zero if $\widetilde{p}_{\theta}(\widetilde{\bx}_0|\bx_t)$ places all of its probability mass on the original value $\bx_0$.
The decomposition of $q(\bx_{t-1}|\bx_{t}, \bx_0)$ in \eqref{eq:discrete_t_step_posterior} also provides us with a motivation for this parameterization. According to  \eqref{eq:discrete_t_step_posterior},  in a given state $\bx_t$, the optimal reverse process only takes into account transitions to states for which $q(\bx_t | \bx_{t-1})$ is non-zero. Therefore, the sparsity pattern of $\bs Q_t$ determines the sparsity pattern of the ideal reverse transition probabilities in $p_{\theta}(\bx_{t-1}| \bx_{t})$. The parameterization in \eqref{eq:x_0parameterization} automatically ensures that the learned reverse probability distribution $p_{\theta}(\bx_{t-1}|\bx_{t})$ has the correct sparsity pattern dictated by the choice of the Markov transition matrix $\bs Q_t$. This parameterization also lets us perform inference with $k$ steps at a time, by predicting $p_\theta(\bx_{t-k} | \bx_t) = \sum q(\bx_{t-k}, \bx_t |  \widetilde{\bx}_0)\widetilde{p_\theta}(\widetilde{\bx}_0 | \bx_t)$.

Finally, when modeling ordinal discrete data, instead of predicting the logits of $\widetilde p_{\theta}(\widetilde{\bx}_0|\bx_t)$ directly with the output of a neural net, another option is to model the probabilities with a truncated discretized logistic distribution (see Appendix~\ref{sec:appendix_logistic}). This provides an extra ordinal inductive bias to the reverse model and boosts FID and log-likelihood scores for images.

\subsection{Loss function}
While the original diffusion models introduced by \citet{sohl2015deep} were optimized with the negative variational lower bound $L_{\mathrm{vb}}$ of \eqref{eq:negative_lower_bound}, more recent diffusion models are optimized with different objectives.
For instance, \citet{ho2020denoising} derive a simplified loss function ($L_{\mathrm{simple}}$) that reweights the negative variational bound, and  \citet{nichol2021improved} explore a hybrid loss $L_{\mathrm{hybrid}} = L_{\mathrm{simple}} + \lambda L_{\mathrm{vb}}$ (using one term to learn the predicted mean and the other to learn predicted variance).
Inspired by this recent work, we introduce an auxiliary denoising objective for the $\bx_0$-parameterization of the reverse process, which encourages good predictions of the data $\bx_0$ at each time step. 
We combine this with the negative variational lower bound, yielding the following alternative loss function:
\begin{align}
    L_{\lambda} =& L_{\mathrm{vb}} + \lambda \; \mathbb E_{q(\bx_0)}\mathbb E_{q(\bx_t|\bx_0)}[-\log \widetilde p_{\theta}(\bx_0|\bx_t)].
    \label{eq:alternative_loss}
\end{align}
Note that the auxiliary loss coincides with the cross entropy term $L_0$ in \eqref{eq:negative_lower_bound} at $t=1$. Furthermore, due to the $\bx_0$-parameterization of $p_{\theta}(\bx_{t-1}|\bx_t)$, both the auxiliary loss term and $D_{\mathrm{KL}}[q(\bx_{t-1} | \bx_t, \bx_0) \vert\vert p_{\theta}(\bx_{t-1}|\bx_t)]$ in $L_{\mathrm{vb}}$ are minimized exactly when $\widetilde p_{\theta}(\widetilde{\bx}_0|\bx_t)$ has all its mass on the datapoint $\bx_0$.
We find that training with this loss leads to improved quality of image samples. 

\section{Connection to existing probabilistic models for text}
\label{sec:bert_connections}
In this section we expand on interesting connections between the D3PM framework and several existing probabilistic and language modeling approaches.

\textbf{BERT is a one-step diffusion model:} One possible D3PM transition matrix is a combination of a uniform transition matrix and an absorbing state at the [MASK] token
(i.e. $\bs{Q} = \alpha \mathbbm{1} e_m^T + \beta \mathbbm{1}\mathbbm{1}^T/K + (1 - \alpha - \beta) I$, where $e_m$ is a one-hot vector on the [MASK] token).
For a one-step diffusion process in which $q(\bx_1 | \bx_0)$ replaces 10\% of tokens with [MASK] and 5\% uniformly at random, this leads precisely to the BERT denoising objective, i.e. $L_{vb} - L_T = - \mathbb E_{q(\bx_1|\bx_0)} [\log p_{\theta}(\bx_0|\bx_1)] = L_{BERT}$,
since $L_T$ is a constant independent of $\theta$ (assuming a fixed prior).

\textbf{Autoregressive models are (discrete) diffusion models:} Consider a diffusion process that deterministically masks tokens one-by-one in a sequence of length $N=T$:
$q(\left[\bx_t\right]_i \mid \bx_0) = [\bx_0]_i \text{ if } i < N - t \text{ else [MASK] }$. This is a deterministic forward process, so $q(\bx_{t-1} | \bx_t, \bx_0)$ is a delta distribution on the $\bx_t$ sequence with one fewer mask: $q(\left[\bx_{t-1}\right]_i | \bx_t, \bx_0) = \delta_{[\bx_{t}]_i} \text{ if } i\neq T - t \text{ else } \delta_{[\bx_0]_i}$. While this process is not applied independently to each token, it can
be recast as an independently-applied diffusion process on the product space
$[0...N]\times\mathcal{V}$, where each token is tagged with its position in the sequence, $\mathcal{V}$ is the vocabulary, and $\bs{Q}$ is an $N\times|\mathcal V|\times N\times |\mathcal V|$ sparse matrix.

Because all tokens except the one at position $i = T - t$ have deterministic posteriors, the KL divergence $D_{KL}(q([\bx_{t-1}]_j | \bx_t, \bx_0)\mid\mid p_\theta([\bx_{t-1}]_j | \bx_t))$ is zero for all other positions. The only token for which this is not true is the token at position $i$, for which $D_{KL}(q([\bx_{t-1}]_i| \bx_t, \bx_0)\mid\mid p_\theta([\bx_{t-1}]_i | \bx_t)) = -\log p_\theta([\bx_0]_i | \bx_t)$, the standard cross entropy loss for an autoregressive model.

\textbf{(Generative) Masked Language-Models (MLMs) are diffusion models:} Generative Masked Language Models (\citep{mask_predict}, \citep{bert_speak}) are generative models that generate text from a sequence of [MASK] tokens. They are usually trained by sampling a sequence $\bx_0$, masking $k$ tokens according to some schedule, and learning to predict the masked tokens given context. It turns out that a D3PM absorbing ([MASK]) model trained on the usual ELBO objective with
% the $\beta(t) = 1 / (T - t + 1)$ schedule and
the $\bx_0$-parameterization from \ref{sec:x0_parameterization} reduces to a reweighted version of this MLM objective (see Appendix \ref{appendix:cmlm} for a detailed derivation).

\section{Text generation}

For text, we experiment with generation on two datasets: text8 \citep{text8}, a character-level dataset extracted from English-language Wikipedia, and the One Billion Word dataset (LM1B) \citep{lm1b}, a large dataset of shuffled English-language sentences. For both, we train a D3PM uniform model based on the work by \citet{hoogeboom2021argmax} (D3PM uniform) and a model that masks tokens (D3PM absorbing). We also consider a model that transitions uniformly to nearest neighbors in a token embedding space (D3PM NN). We follow \citet{hoogeboom2021argmax} and use $T=1000$ timesteps, although we are also able to evaluate on fewer due to the parameterization in Section \ref{sec:x0_parameterization}.

\subsection{Character-level generation on text8}
\begin{table}[b]
  \caption{Quantitative results on text8. NLL is reported on the entire test set. Sample times are for generating a single example of length 256. Results are reported on two seeds. All models are standard 12-layer transformers unless otherwise noted. $^\dagger$Transformer XL is a 24-layer transformer, using a 784 context window. $^\ddagger$Results reported by \citep{hoogeboom2021argmax} by running code from official repository.}
  \label{tab:text8}
  \scriptsize
  \centering
  \begin{tabular}{llll}
    \toprule
    % \multicolumn{2}{c}{Part}                   \\
    % \cmidrule(r){1-2}
    Model & Model steps & NLL (bits/char) ($\downarrow$) & Sample time (s) ($\downarrow$)\\
    \midrule
    Discrete Flow \citep{tran2019discrete} ($8\times 3$ layers) & - & $1.23$ & $0.16$ \\
    Argmax Coupling Flow \citep{hoogeboom2021argmax} & - & $1.80$ & $0.40 \pm 0.03$ \\
    IAF / SCF \citep{Ziegler2019}$^\ddagger$ & - & $1.88$ & $0.04 \pm 0.0004$\\
    Multinomial Diffusion (D3PM uniform) \citep{hoogeboom2021argmax} & $1000$  & $\leq 1.72$ & $26.6\pm 2.2$   \\
    \midrule
    D3PM uniform \citep{hoogeboom2021argmax} (ours) & $1000$  & $\leq 1.61\pm 0.02$ & $3.6\pm 0.4$  \\
    D3PM NN ($L_{\mathrm{vb}}$) (ours) & $1000$ & $\leq 1.59\pm 0.03$ & $3.1474\pm 0.0002$ \\
    D3PM mask ($L_{\lambda=0.01}$) (ours) & $1000$ & $\leq 1.45\pm 0.02$ & $3.4\pm 0.3$ \\
    \midrule
    D3PM uniform \citep{hoogeboom2021argmax} (ours) & $256$ & $\leq 1.68\pm0.01$ & $0.5801\pm 0.0001$ \\
    D3PM NN ($L_{\mathrm{vb}}$) (ours) & $256$ & $\leq 1.64\pm 0.02$ & $0.813\pm 0.002$ \\
    D3PM absorbing ($L_{\lambda=0.01}$) (ours) & $256$ & $\leq 1.47\pm 0.03$ & $0.598\pm 0.002$ \\
    Transformer decoder (ours) & $256$ & $1.23$ & $0.3570\pm 0.0002$\\
    Transformer decoder \citep{al_rfou} & $256$ & $1.18$ & -\\
    Transformer XL \citep{transformer_xl}$^\dagger$ & $256$ & $1.08$ & - \\
    \midrule
    D3PM uniform \citep{hoogeboom2021argmax} (ours) & $20$ & $\leq1.79\pm 0.03$ & $0.0771\pm 0.0005$ \\   
    D3PM NN ($L_{\mathrm{vb}}$) (ours) & $20$ & $\leq 1.75\pm0.02$ & $0.1110\pm 0.0001$ \\
    D3PM absorbing ($L_{\lambda=0.01}$) (ours) & $20$ & $\leq 1.56\pm0.04$ & $0.0785\pm 0.0003$ \\
    \bottomrule
  \end{tabular}
\end{table}

text8 is a character-level text dataset consisting of a small vocabulary of 27 tokens: the letters `a'-`z' and the `\_' whitespace token. We follow the convention of training and evaluating text8 in chunks of length 256 without any preprocessing \citep{hoogeboom2021argmax}. For nearest-neighbor D3PM, our nearest neighbor graph in character-space is shown in Appendix \ref{appendix:text8_aux}. D3PM uniform models were trained with a cosine schedule from \citet{hoogeboom2021argmax} (ablations in Appendix~\ref{appendix:text8_aux}), while D3PM absorbing and D3PM NN models were trained with a mutual information schedule.

Table \ref{tab:text8} shows that for D3PM, the D3PM absorbing model performed the best, exceeding the uniform and NN diffusion models. We were able to improve upon the baseline result of \citep{hoogeboom2021argmax} with hyperparameter tuning, and our uniform and NN results outperformed results from \citet{hoogeboom2021argmax} across all inference steps, down to as few as 20. We found that $L_{\lambda=0.01}$ worked best for D3PM absorbing, while $L_{\mathrm{vb}}$ was better for D3PM uniform. Our model outperforms all non-autoregressive baselines except one, the Discrete Flow model \citep{tran2019discrete} 
(for which unfortunately no open-source implementations exist), and is also faster than all but one method, the IAF/SCF model \citep{Ziegler2019}. It is also nearly 20x faster than an autoregressive transformer of the same size. We also include a plot of inference time as a function of iterations in Appendix~\ref{appendix:text8_aux}.
D3PM with the mask absorbing token was by far the best performing model, which lends credibility to the use of masks in denoising auto-encoders. Nearest-neighbor diffusion only narrowly improves upon a D3PM-uniform model: this was a surprising negative result for us, suggesting that not all notions of structure are meaningful.

\subsection{Text generation on LM1B}
Text generation for large-scale text datasets and large vocabularies with discrete diffusion models has not been previously demonstrated. We include results from LM1B as a proof of concept, showing that these models can indeed scale (as discussed in Appendix \ref{sec:scaling-to-large-categories}), and that the D3PM absorbing model continues to excel. All models were trained and evaluated on packed sequences of length $128$, using a sentencepiece\footnote{\url{https://github.com/google/sentencepiece}} vocabulary of size $8192$.
\begin{figure}[t]
    \centering
    \includegraphics[width=\textwidth]{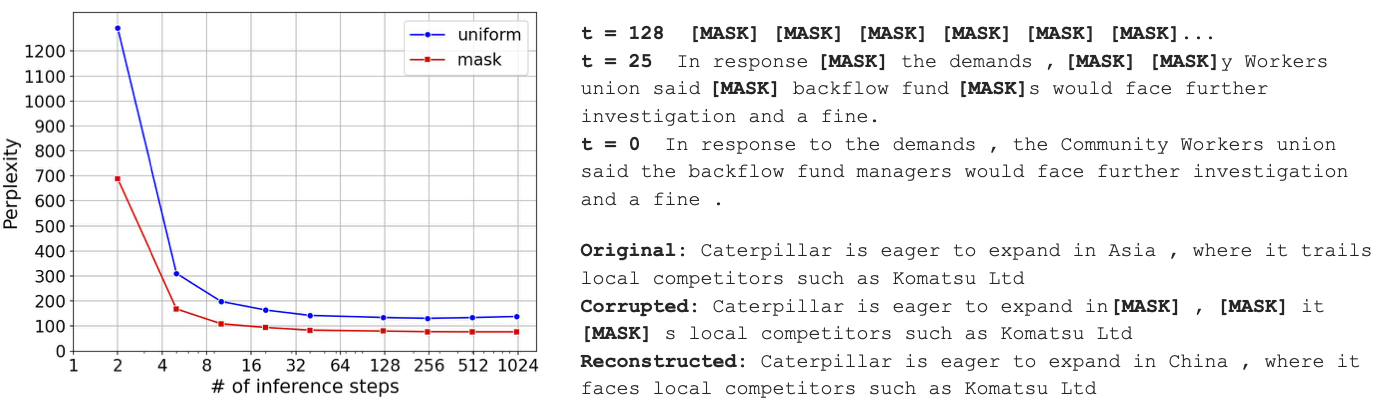}
    \caption{Left: perplexity v.s. sampling iterations for LM1B. Right: Using a trained D3PM absorbing model for LM1B to (top) generate new sentences and (bottom) reconstruct corrupted examples.}
    \vspace{-0.3cm}
    \label{fig:lm1b_scaling}
\end{figure}

Table \ref{tab:lm1b} contains results from experiments on LM1B. Overall, mask diffusion (D3PM absorbing) does relatively well, approaching the performance of a comparable autoregressive model of the same size, and scaling to far fewer steps, while uniform diffusion performs significantly worse. We find, surprisingly, that the D3PM NN model performs worse than the uniform model in terms of log likelihoods (although it demonstrates unique qualitative behavior). This suggests that word embedding similarity may not be a meaningful kind of locality in a diffusion process. We found the the $L_{\lambda=0.01}$ loss worked best for the mask absorbing model, but reduced performance for the other models. We note the surprising scaling in perplexity in Figure \ref{fig:lm1b_scaling}, achieving strong results with as few as 10 inference steps. We also show samples from our model and completions from corrupted samples.
\begin{table}[t] % Daniel: Strong opinion against inline tables. Top or bottom only.
\vspace{-0.3cm}
  \caption{Quantitative results on LM1B. Perplexity reported on the test set. Results are reported on two seeds. All models have context window length 128 and 12 layers unless otherwise noted. $^\dagger$Transformer XL is a 24 layer transformer. $^\ddagger$rounded for readability, see Appendix~\ref{appendix:lm1b_aux}.}
  \label{tab:lm1b}
  \scriptsize
  \centering
  \begin{tabular}{r rrr c rrr}
    \toprule
    Metric: &\multicolumn{3}{c}{Perplexity ($\downarrow$)}
    &&\multicolumn{3}{c}{Sample time$^\ddagger$ (s) ($\downarrow$)}\\
    \cmidrule{2-4}\cmidrule{6-8}
    inference steps: & $1000$
    & $128$
    & $64$
    &
    & $1000$
    & $128$
    & $64$\\
    \midrule
    D3PM uniform
    & 137.9 $\pm$ 2.1 & 139.2 $\pm$ 1.2 & 145.0 $\pm$ 1.2 &
    & 1.82 & 0.21 & 0.08 \\
    D3PM NN
    & 149.5 $\pm$ 1.3 & 158.6 $\pm$ 2.2 & 160.4 $\pm$ 1.2 &
    & 21.29 & 6.69 & 5.88 \\
    D3PM absorbing
    & 76.9 $\pm$ 2.3 & 80.1 $\pm$ 1.2 & 83.6 $\pm$ 6.1 &
    & 1.90 & 0.19 & 0.10 \\
    \midrule
    Transformer (ours)
    & - & 43.6 & - &
    & - & 0.26 & -\\
    Transformer XL \citep{transformer_xl}$^\dagger$
    & - & 21.8 & - &
    & - & - & -\\
    \bottomrule
  \end{tabular}
%   \begin{tabular}{lllll}
%     \toprule
%     % \multicolumn{2}{c}{Part}                   \\
%     % \cmidrule(r){1-2}
%     Model & Perplexity @ 1000 steps & @ 128 steps & @ 64 steps & Sample time (at 1000/128/64)\\
%     \midrule
%     D3PM uniform \citep{hoogeboom2021argmax} (ours) & $159.85$ & $165.93$ & - & a/b/c \\
%     D3PM mask ($L_{\lambda=0.01}$) (ours) & $78.15\pm 4.1$ & $83.172$ & $89.13\pm 9.2$ & a/b/c \\
%     \midrule
%     Transformer decoder (ours) & - & $43.62549$ & - & -/b/- \\
%     Transformer XL \citep{transformer_xl}$^\dagger$ & - & $21.8$ & - & -/c/- \\
%     \bottomrule
%   \end{tabular}
\end{table}

\section{Image generation}

\begin{table}[t]
  \caption{Inception scores (IS), Frechet Inception Distance (FID) and negative log-likehood (NLL) on the image dataset CIFAR-10.  The NLL is reported on the test set in bits per dimension. We report our results as averages with standard deviations, obtained by training five models with different seeds.}
  \label{tab:cifar10}
  \scriptsize
  \centering
  \begin{tabular}{llll}
    \toprule
    \multicolumn{1}{l}{Model}  & IS ($\uparrow$)    & FID ($\downarrow$) & NLL ($\downarrow$)        \\
    \midrule
    Sparse Transformer \citep{child2019generating}    &        &       & 2.80\\
    NCSN  \citep{song_2019}                 & $8.87\pm0.12$ & 25.32  &\\
    NCSNv2  \citep{song2020improved}        & $8.40\pm0.07$ & 10.87  &\\
    StyleGAN2 + ADA  \citep{karras2020training}      & $9.74\pm0.05$ & 3.26 &\\
    \midrule
    Diffusion (original), $L_{\mathrm{vb}}$ \citep{sohl2015deep}  & & & $\leq 5.40$ \\
    DDPM $L_{\mathrm{vb}}$ \citep{ho2020denoising}  & $7.67 \pm 0.13$  &  $13.51$ & $\leq 3.70$  \\
    DDPM $L_{\mathrm{simple}}$ \citep{ho2020denoising} & $9.46 \pm 0.11$  &  $3.17$ & $\leq 3.75$ \\
    Improved DDPM $L_{\mathrm{vb}}$ \citep{nichol2021improved} &   & $11.47$ & $\leq 2.94$ \\
    Improved DDPM $L_{\mathrm{simple}}$ \citep{nichol2021improved} &   & $2.90$ & $\leq 3.37$ \\
    DDPM++ cont \citep{song2020_sde} &   &  $2.92$ & $2.99$ \\
    NCSN++ cont. \citep{song2020_sde}  & $9.89$ & $2.20$ & \\
    \midrule
    D3PM uniform $L_{\mathrm{vb}}$ & $5.99 \pm 0.14$  &  $51.27 \pm 2.15$ & $\leq 5.08 \pm 0.02$  \\
    D3PM absorbing $L_{\mathrm{vb}}$ & $6.26 \pm 0.10$  &  $41.28 \pm 0.65$ & $\leq 4.83 \pm 0.02$  \\
    D3PM absorbing $L_{\lambda=0.001}$  & $6.78 \pm 0.08$  &  $30.97 \pm 0.64$ & $\leq 4.40 \pm 0.02$ \\
    D3PM Gauss $L_{\mathrm{vb}}$ & $7.75 \pm 0.13$  &  $15.30 \pm 0.55$ & $\leq 3.966 \pm 0.005$  \\
    D3PM Gauss  $L_{\lambda=0.001}$ & $8.54 \pm 0.12$  &  $8.34 \pm 0.10$ & $\leq 3.975 \pm 0.006$ \\
    D3PM Gauss + logistic $L_{\lambda=0.001}$ & $8.56 \pm 0.10$  &  $7.34 \pm 0.19$ &  $\leq 3.435 \pm 0.007$ \\
    \bottomrule
  \end{tabular}
\end{table}

We evaluate the performance of several D3PM models on the task of unconditional image generation with the dataset CIFAR-10 \citep{krizhevsky2009learning}. We follow \citet{ho2020denoising} and use $T=1000$ timesteps for all models and verify that for all models the forward process converges to the stationary distribution within $T$ steps, yielding a value of at most $L_T\approx 10^{-5}$ bits per dimension. We train three versions of D3PM with different transition matrices: doubly stochastic matrices with uniform transition probabilities (D3PM uniform) \citep{hoogeboom2021argmax}, transition matrices with an absorbing state located at R, G and B values of 128 (D3PM absorbing) and doubly stochastic discretized Gaussian transition matrices (D3PM Gauss). For the D3PM uniform model we experimented with a linear $\beta_t$ schedule as well as the cosine schedule as proposed in \citep{hoogeboom2021argmax}, with the cosine schedule producing the best results. For D3PM absorbing we use the schedule $\beta_t = (T-t+1)^{-1}$ as also proposed in \citep{sohl2015deep}, which corresponds to increasing the probability of being in the absorbing state linearly over time.
For D3PM Gauss we use the same linear schedule as in \citep{ho2020denoising}. See Appendix \ref{sec:appendix_image} for more details on the experimental setup.

\begin{figure}[b]
    \centering
    \begin{subfigure}[b]{0.54\textwidth}
         \centering
         \includegraphics[width=\textwidth]{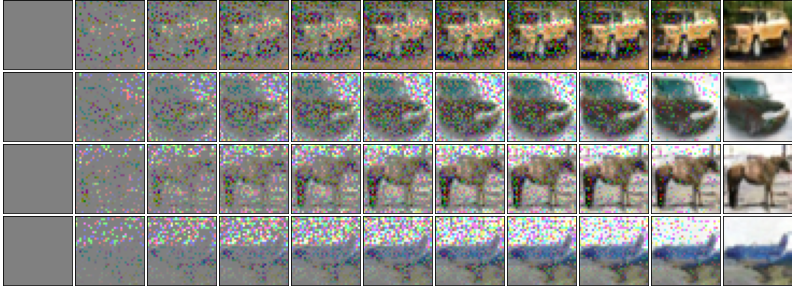} \\
         \vspace{0.07cm}
         \includegraphics[width=\textwidth]{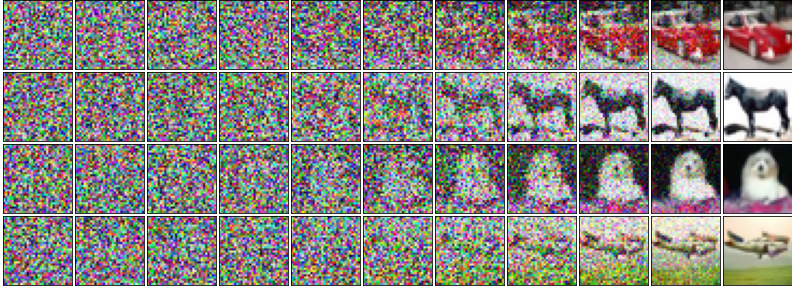} 
     \end{subfigure}
     \hfill
     \begin{subfigure}[b]{0.45\textwidth}
         \centering
         \includegraphics[width=\textwidth]{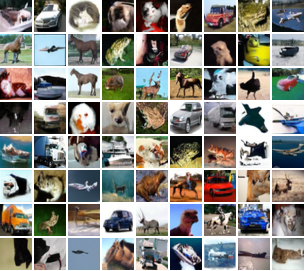}
     \end{subfigure}
    \caption{Left: progressive sampling at $t=1000, 900, 800, ..., 0$ for D3PM absorbing (top) and D3PM Gauss + logistic (bottom), trained with $L_{\lambda}$ loss on CIFAR-10. These samples were cherry picked. Right: (non cherry picked) samples from the D3PM Gauss + logistic model.}
    \label{fig:cifar10_samples}
\end{figure}
Table~\ref{tab:cifar10} shows that for D3PM models trained with the $L_{\mathrm{vb}}$ objective, D3PM Gauss performs better than D3PM absorbing and uniform on all metrics: Inception score (IS), Frechet Inception Distance (FID) and negative log-likelihood (NLL). The IS score of the uniform and absorbing D3PM models are comparable, while the FID score and NLL of the D3PM absorbing model are slightly better. We trained both D3PM absorbing and D3PM Gauss with the alternative loss function $L_{\lambda}$ of \eqref{eq:alternative_loss}, and we found $\lambda=0.001$ to work best.
We have also experimented with larger values of $\lambda$ and a model trained only with the auxiliary denoising term in \eqref{eq:alternative_loss}. Although this led to a more rapid increase in performance early on in training, the NLL leveled off at higher values for larger $\lambda$ and the FID even started increasing again.
%}
The results show that the models trained with $L_{\lambda}$ perform significantly better than their counterparts trained with $L_{\mathrm{vb}}$.
One explanation for this boost in performance is that the cross entropy term leads to gradient noise that varies less with the time step $t$, which is in contrast to the large change in magnitude of the $L_{t-1}$ terms in $L_{\mathrm{vb}}$ for smaller $t$, as demonstrated by \citet{nichol2021improved}. 
Finally, we achieve our best results by combining D3PM Gauss trained on $L_{\lambda}$ with a truncated logistic parameterization of the reverse process distribution $p_{\theta}(\widetilde{\bx}_0|\bx_t)$ (D3PM Gauss + logistic). Figure~\ref{fig:cifar10_samples} shows samples from our best model (D3PM Gauss + logistic), as well as the D3PM absorbing model.

\section{Related Work}

Diffusion generative models were first proposed by \citet{sohl2015deep} and have gained renewed attention recently due to strong results on image and waveform generation \citep{ho2020denoising,wavegrad}. Recent works have proposed improvements for diffusion model training, including importance sampling of the ELBO, better noise schedules \citep{nichol2021improved} and implicit diffusion models \citep{song2021implicit}. Several works have also drawn connections to score matching \citep{vincent2011connection,hyvarinen2004independent,song_2019}, leading to improved sampling algorithms in the continuous-time limit \citep{song2020_sde}.

While most works have considered continuous diffusion models, discrete diffusion-like models were described in \citep{sohl2015deep} and applied to text generation and image segmentation data in \citep{hoogeboom2021argmax}. Some works \citep{permutation,music} have dealt with discrete data by embedding it in continuous space and leveraging Gaussian diffusion, but have not applied this to text. \citet{princeton_chem} also considered generation of discrete structured objects using a diffusion-like Markov corruption process.

For text, denoising autoencoders have a long history both in representation learning \citep{Bengio_2013, bert} and more recently as generative models \citep{bert_speak}. These closely resemble our absorbing state diffusion variants for a particular schedule and transition matrix (see Section \ref{sec:bert_connections}), although our framing allows us to compute log-likelihoods and experiment with alternative transition matrices. Other works have considered non-autoregressive translation and speech transcription via insertion and deletion \citep{levenstein,ruis2020insertion}, masking \citep{mask_predict}, and iteratively-refined sequence alignments \citep{chan2020imputer,saharia2020latentalignments}.

\section{Discussion}
\label{sec:discussion}
We have presented D3PMs, a class of models that improves diffusion models for discrete data by defining new kinds of discrete corruption processes.
We achieve strong empirical results relative to previous work on discrete diffusion models, even surpassing performance of continuous diffusion models in terms of log-likelihoods for image generation.
While these results are promising, one limitation is that---like much other work on non-autoregressive generative models---our models are still inferior to strong autoregressive models like Transformer XL for text generation, and continuous diffusion models still yield stronger results on image quality.
We expect that D3PMs can benefit further from the rapid development of continuous diffusion models \citep{song2020_sde, nichol2021improved}. For example, further research in alternative losses for D3PM's can take inspiration from the reweighted $L_{\mathrm{simple}}$ objective used in \citep{ho2020denoising}, or the resampled variational bound in \citet{nichol2021improved}. Furthermore, D3PM's might benefit from increasing the number of timesteps and a more optimized noise schedule, as discussed in \citet{nichol2021improved}.
Another limitation comes from the choice of evaluation metrics that we use (and that are standard for evaluation of generative models). 
Inception score and Frechet Inception Distance are based on neural networks that have been trained on a particular distribution of data, which is not representative for all use-cases, and focusing on average quality metrics may not accurately reflect performance across the wide diversity of settings where these generative models may be applied.
This creates a risk of negative social impacts where advances disproportionately favor a subset of the population.
Going forward, we are excited about the space of possibilities that arise within the D3PM framework. 
We have found successes in leveraging the flexibility that comes from defining discrete corruption processes for discrete data, but we believe that there are many more possibilities that make use of richer forms of structure to define even more powerful discrete diffusion models.

\begin{ack}
We would like to thank Hugo Larochelle for providing high-level feedback during the project, and Ben Poole for reviewing a draft version of this manuscript. We would also like to thank Julia Kreutzer and Xavier Garcia for helpful conversations about language experiments. We, the authors, declare to have no competing interests. The research conducted for this paper was entirely supported by Google. 
\end{ack}

\bibliographystyle{plainnat}
\bibliography{main}

%%%%%%%%%%%%%%%%%%%%%%%%%%%%%%%%%%%%%%%%%%%%%%%%%%%%%%%%%%%%

     % \include{sections/checklist}
     \appendix
\newpage
\section{Additional details regarding D3PMs}
%%%%%%%%%%%%%%%%%%%%%%%%%%%%%%%%%%%%%%%%%%%%%%%%%%%%%%%%%%%%%%%%%%%%%%%%%%%%%%%%%%%%%%%%%%%%%%%%
\subsection{Doubly-stochastic matrices}
\label{appendix:stochastic}

As discussed in Section \ref{sec:choice-transition-matrices}, there are two constraints on $\bs{Q}_t$ that allow it to be used within a D3PM: the rows of $\bs Q_t$ must sum to one to conserve probability mass, and the rows of $\overline{\bs Q}_{t} = \bs Q_1  \bs Q_2 \hdots \bs Q_t$ must converge to a known stationary distribution as $t$ becomes large. Technically, it is also possible to use a learned prior $p_\theta(\bx_T)$, but assuming this is still modeled under a conditional independence assumption, $q(\bx_T | \bx_0)$ must still be close to a stationary distribution for the $L_T$ loss term to be small.

One way to ensure that this occurs is to chose $\bs{Q}_t$ as increasing powers of a doubly stochastic base matrix $\bQ$ (rows and columns sum to 1) with strictly positive entries. This is enough to ensure that $\bQ$ is is irreducible and aperiodic and that product $\overline{\bs Q}_t$ converges as $t\rightarrow\infty$ to a uniform distribution over all states.
To show this, consider $\pi_i = 1/K$ for $i=1, ..., K$, and $\sum_{i=1}^K \bQ_{i,:} = \bs 1$ and $\sum_{j=1}^K \bQ_{:, j} = \bs 1$, then $[\bs Q \bs \pi]_i = \sum_{j=1}^K \bQ_{i, j} \pi_j =1/K\sum_{j=1}^K \bQ_{i, j} = 1/K = \pi_i$, thus the uniform distribution is an eigenvector of the transition matrix with eigenvalue 1. Convergence to this distribution follows from the Perron-Frobenius theorem for positive square matrices.

More generally, a similar argument shows that even for $\bQ_t$ that are not powers of the same base matrix, as long as each $\bQ_t$ is doubly stochastic, irreducible, and aperiodic, the uniform distribution is the only possible stationary distribution, and as long as the second largest eigenvalue of $\bQ_t$ is bounded below, the cumulative product $\barbQ_t$ will converge to the uniform distribution. In practice, we choose $\bQ_t$ to add more noise as $t$ increases, which ensures that $\overline{\bs Q}_T$ is very close to reaching a uniform stationary distribution.

%%%%%%%%%%%%%%%%%%%%%%%%%%%%%%%%%%%%%%%%%%%%%%%%%%%%%%%%%%%%%%%%%%%%%%%%%%%%%%%%%%%%%%%%%%%%%%%%
\subsection{More details on possible choices of Markov transition matrices}
\label{sec:appendix_other_transition_mats}

\subsubsection{Uniform diffusion}
\label{sec:appendix_transition_uniform}

The transition matrix described by \citet{sohl2015deep} for the binary case, and extended by \citet{hoogeboom2021argmax}, to the categorical case, can be represented using the following $K\times K$ transition matrix
\begin{align}
    \left[\bs{Q}_t\right]_{ij} = \begin{cases}
    1 - \frac{K-1}{K} \beta_t \quad &\text{if}\quad i=j\\
    \frac{1}{K} \beta_t \quad &\text{if}\quad i\neq j
    \end{cases},
    \label{eq:cat_forward_kernel_mat_uniform}
\end{align} 
This transition matrix can also be written as $(1 - \beta_t) I + \beta_t \mathbbm{1} \mathbbm{1}^T / K$, where $\mathbbm{1}$ is a column vector of all ones.

\subsubsection{Diffusion with an absorbing state}
\label{sec:appendix_transition_mask}
For our diffusion models with an absorbing state $m$, we use the following matrix:

\begin{align}
    \left[\bs{Q}_t\right]_{ij} = \begin{cases}
    1  \quad&\text{if}  \quad i = j = m \\
    1 - \beta_t \quad &\text{if} \quad i = j \ne m \\
    \beta_t \quad &\text{if} \quad j = m, i \ne m\\
\end{cases}
    \label{eq:cat_forward_kernel_mat_mask}
\end{align}

The transition matrix can also be written as $(1 - \beta_t) I + \beta_t \mathbbm{1} e^T_m$, where $e_m$ is a vector with a one on the absorbing state $m$ and zeros elsewhere.
Since $m$ is an absorbing state, the corruption process converges not to a uniform distribution but to the point-mass distribution on $m$. 

For text generation, we let $m$ be the [MASK] token at index $K-1$; this leads to a BERT-like training objective, which masks tokens according to some schedule and learns to denoise them iteratively (see Section \ref{sec:bert_connections}). For image generation, we set $m$ to the gray RGB pixel $(128, 128, 128)$ at index $K//2$.

\subsubsection{Discretized Gaussian transition matrices}
\label{sec:appendix_transition_gaussian}

For our D3PM models applied to ordinal data, inspired by continuous-space diffusion models, we use the following $K \times K$ matrix:
\begin{align}
    \left[\bs{Q}_t\right]_{ij} = \begin{cases}
\frac{\exp\left(-\frac{4|i - j|^2 }{(K-1)^2\beta_t}\right)}{\sum_{n=-(K-1)}^{K-1}\exp\left(-\frac{4n^2 }{(K-1)^2\beta_t}\right)} \quad &\text{if} \quad i\neq j\\[1.5em]
1 - \sum_{l=0, l\neq i}^{K-1} [\bs Q_t]_{il} \quad &\text{if} \quad i = j \\
\end{cases}
    \label{eq:discretized_gaussian_mat}
\end{align}

Normalization is ensured by assigning the diagonal values to one minus the sum of each row (not including the diagonal entry). Note that due to the normalization of the off-diagonal values over the range $\{-K+1, ..., K-1\}$ the sum of each row excluding the diagonal entry is always smaller than 1. The result yields an irreducible doubly stochastic matrix and a forward process with a uniform stationary distribution. Similar to the continuous Gaussian diffusion model, the parameters $\beta_t$ influence the variance of the forward process distributions.

\subsubsection{Structured diffusion in text: using word-embedding distance to introduce locality}\label{sec:appendix_transition_nearest_neighbor}

For text, we construct a $k$-nearest neighbor adjacency matrix \[
[\mathbf{G}]_{ij} = 1\text{ if } w_i \text{ is a k-nearest neighbor of } w_j \text{ else } 0
\]
constructed from a pre-trained embedding space over the vocabulary. Then we consider a symmetrized adjacency matrix of the form $\mathbf{A}=(\mathbf{G} + \mathbf{G}^T) / (2k)$ where $k$ is the number of nearest neighbors of each node,
and finally construct a doubly stochastic rate matrix with
\begin{align}
    \left[\bs{R}\right]_{ij} = \begin{cases}
        - \sum_{l\neq i} A_{il} \quad &\text{if} \quad i = j \\
        A_{ij} \quad &\text{ otherwise }\\
    \end{cases}
    \label{eq:nn_diffusion}
\end{align}
Our final transition matrix is constructed as a matrix exponential of this rate matrix:
$$
\mathbf{Q}_t = \exp(\alpha_t \mathbf{R}) = \sum_{n = 0}^\infty \frac{\alpha_t^n}{n!} \bs{R}^n
$$
Since $\bs{R}$ is symmetric and sums to zero along each row, $\mathbf{Q}_t$ is doubly stochastic, which ensures we have a uniform stationary distribution (as long as $G$ is connected). Increasing $\alpha_t$ over time allows us to add more noise for larger values of $t$.

Assuming word embeddings are some metric for syntactic or semantic similarity, this results in a corruption process that gradually moves away from the ground-truth sentence, swapping words with nearest-neighbors in embedding space. For character level modeling, this is a graph over characters, which more often transitions for instance from vowels to other vowels than from vowels to consonants. For words, this could transition between semantically similar words.

\begin{figure}[h]
    \centering
    \includegraphics[scale=0.5]{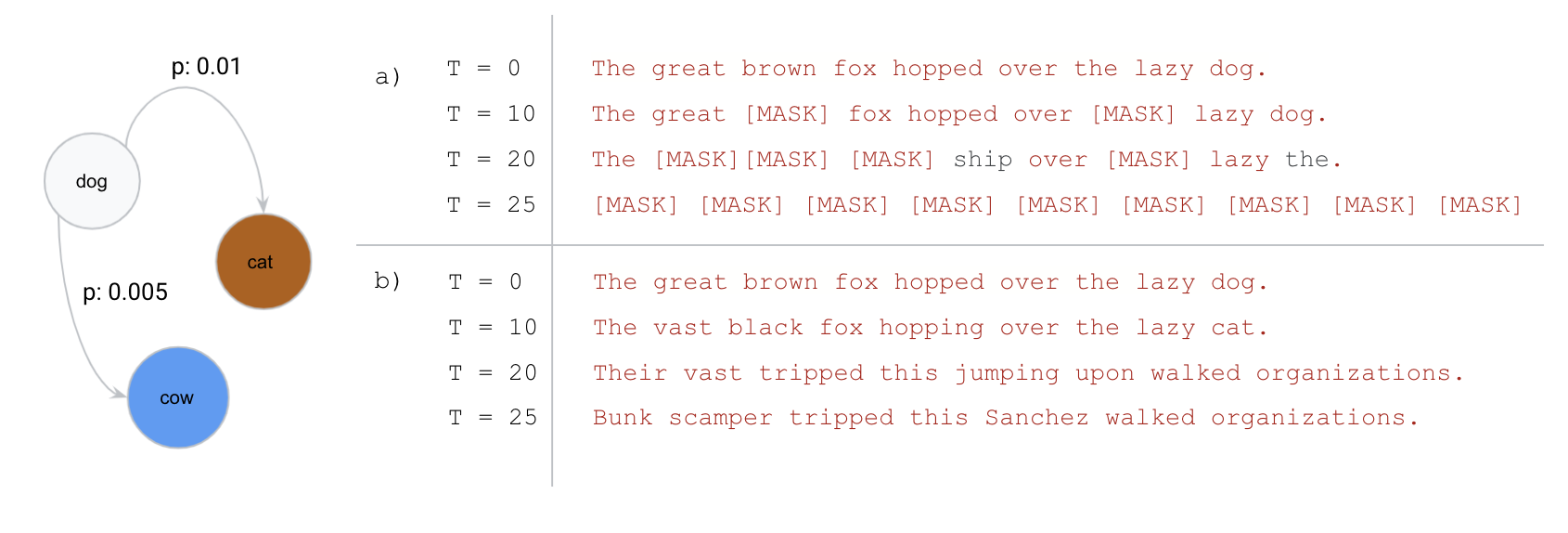}
    \caption{Two examples of noise schedules transforming text data. The top is a BERT-like absorbing + uniform diffusion which replaces tokens with [MASK] tokens (and occasionally with any other token, in black). The bottom is nearest-neighbor diffusion in embedding space. At left represents a possible column in the transition matrix.}
    \label{fig:nn_graph}
\end{figure}

\begin{figure}[h]
    \centering
    \includegraphics[scale=0.5]{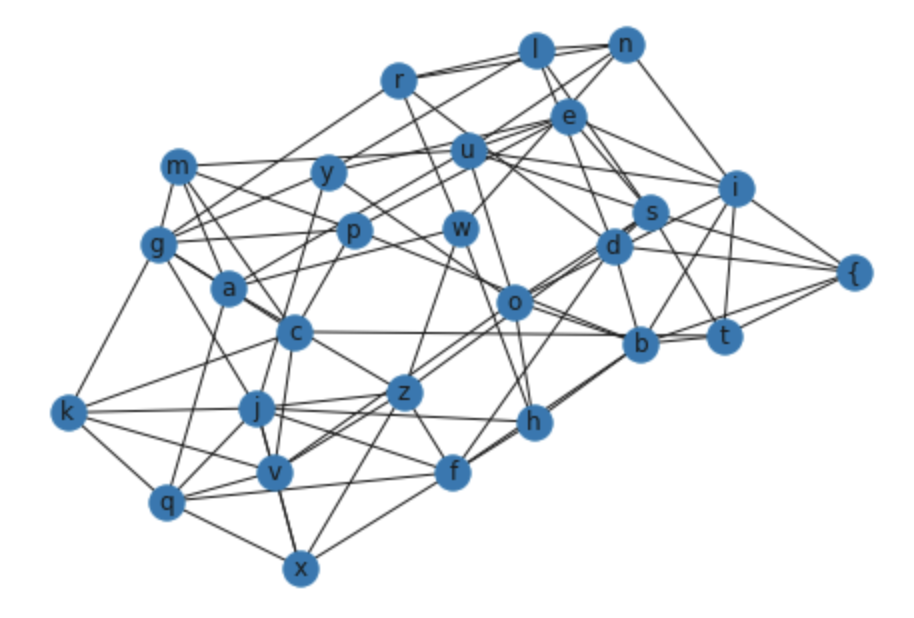}
    \caption{The character-level symmetrized 5-NN graph.}
    \label{fig:char_graph}
\end{figure}

For example, in Figure \ref{fig:nn_graph}, we construct the forward process to diffuse from "dog" to "cat" or "cow", which are nearby in embedding space, but not to more distant words. We can either bootstrap this process by updating the transition matrix $\bs{Q}$ dynamically during training, or use pretrained embeddings; we use pretrained embeddings for all of our experiments.

\subsubsection{Band-diagonal transitions} 
A class of transition matrices that introduce local, ordinal inductive biases for structured data are band-diagonal transition matrices which only allow the corruption process to transition locally between states and biases the reverse process towards local iterative refinement. For example, in images, this can be used to allow transitions only between adjacent pixel values.

\begin{align}
    \left[\bs{Q}_t\right]_{ij} = \begin{cases}
\frac{1}{K} \beta_t \quad &\text{if} \quad  0 < \vert i-j\vert \leq v \\
1 - \sum_{l\neq i} Q_{il} \quad &\text{if} \quad i = j \\
\end{cases}
    \label{eq:uniform_band_diagonal}
\end{align}

where $v$ is the number of nonzero off-diagonal elements of $\bs Q$ above (and below) the main diagonal. Note that this is a doubly stochastic matrix, so the stationary distribution is uniform. We do not use these in our experiments.

\subsubsection{Combinations of absorbing diffusion and other diffusion} 
A few ablations in Appendix~\ref{appendix:text8_aux} consider transition matrices that combine absorbing-state or nearest-neighbor and uniform D3PM models. For instance, an absorbing-uniform transition matrix can be constructed $\bs{Q} = \alpha \mathbbm{1} e_m^T + \beta \mathbbm{1}\mathbbm{1}^T/K + (1 - \alpha - \beta) I$, where $e_m$ is a one-hot vector on the [MASK] token.

%%%%%%%%%%%%%%%%%%%%%%%%%%%%%%%%%%%%%%%%%%%%%%%%%%%%%%%%%%%%%%%%%%%%%%%%%%%%%%%%%%%%%%%%%%%%%%%%
\subsection{Generative Masked Language Models are Diffusion Models}

\label{appendix:cmlm}

Generative Masked Language Models \citep{mask_predict, bert_speak} are generative models that generate text from a sequence of [MASK] tokens. These are usually trained by sampling a sequence $\bx_0$, masking tokens according to some schedule, and learning to predict the masked tokens given context. The actual masking procedure can either be done independently, i.e. by masking each token with probability $p = k/T$, like \citet{bert}, or by sampling exactly $k$ tokens. The usual objective is\footnote{Sometimes the loss is un-normalized or normalized by the full sequence length.}:
\begin{equation}
\label{eq:cmlm}
\min - \mathbb{E}_{q(\bx_0)}\left[\mathbb{E}_{k\in[1...|\bx_0|]}\left[\frac{1}{k}\ \mathbb{E}_{\bx_k \text{with $k$ masked tokens}}\left[\sum_{i \text{ with } [\bx_k]_i = m} \log p_\theta([\bx_0]_i | \bx_k)\right]\right]\right]
\end{equation}
where we first sample a datapoint $\bx_0$, sample a number of tokens to mask $k$ (either uniformly or according to some schedule), then mask that many tokens at random and compute a cross entropy loss over those masked tokens. We claim that this training objective is a (reweighted) absorbing-state D3PM objective with a particular noise schedule and the $\bx_0$-parameterization from \ref{sec:x0_parameterization} (and indeed, that any absorbing-state D3PM model with [MASK] as the absorbing state will be a reweighted version of this loss with different weights assigned to different numbers of masked tokens $k$).

Consider a D3PM with a schedule that masks tokens with probability $\beta_t$. The reverse process predicts $\widetilde{p}_\theta(\widetilde{\bx_0}| \bx_t)$, then uses the forward process to compute $p_{\theta}(\bx_{t-1}|\bx_t) \propto \sum q(\bx_{t-1}, \bx_t| \widetilde{\bx_0}) \widetilde{p}_\theta(\widetilde{\bx}_0| \bx_t)$. In the particular case of absorbing-state diffusion, for each masked token $[\bx_t]_i = m$ in $\bx_t$, we thus have
\[
p_{\theta}([\bx_{t-1}]_i|\bx_t) \propto 
\begin{cases}
[\beta_t \prod_{s < t} (1-\beta_s)] \widetilde{p}_\theta([\widetilde{\bx}_0]_i=[\bx_{0}]_i | \bx_t)
&\text{for }[\bx_{t-1}]_i = [\bx_{0}]_i \ne m
\\
1 - \prod_{s \le t} (1-\beta_s)
&\text{for }[\bx_{t-1}]_i = m
\end{cases}
\]
We note that for each unmasked token $[\bx_t]_i = [\bx_0]_i$, the KL-divergence is zero since unmasked tokens cannot make any other type of transition other than becoming masked. Also, the term in the KL divergence due to the probability of mask transitions is a constant, since mask transitions are independent of the model parameters $\theta$.
Our $L_t$ term is then
$$
D_{\mathrm{KL}}[q(\bx_{t-1} | \bx_t, \bx_0) \vert\vert p_{\theta}(\bx_{t-1}|\bx_t)] = -\left[\beta_t \prod_{s < t} (1-\beta_s)\right] \sum_{i \text{ with } [\bx_t]_i = m} \log \widetilde{p}_\theta([\bx_0]_i | \bx_t) + C
$$
where $C$ is independent of $\theta$ and the sum is taken over the masked tokens in $\bx_t$.
For example, if we use $\beta(t) = 1 / (T - t + 1)$ from \citet{sohl2015deep}, $\beta_t \prod_{i=0}^{t-1} (1 - \beta_i) = 1 / T$ and $1 - \prod_{i=0}^t (1 - \beta_i) = (t-1) / T$, so $q([\bx_{t-1}]_i = [\bx_0]_i | [\bx_{t}]_i = m, \bx_0) = 1 / t$ for non-mask tokens and we can simplify our $L_t$ objective to
$$D_{\mathrm{KL}}[q(\bx_{t-1} | \bx_t, \bx_0) \vert\vert p_{\theta}(\bx_{t-1}|\bx_t)] = -\left[\frac{1}{t}\sum_{i \text{ with } [\bx_t]_i = m} \log \widetilde{p}_\theta([\bx_0]_i | \bx_t)\right] + C$$

where $\bx_t$ masks tokens independently and uniformly with probability $t / T$. The $L_T$ term in our ELBO is 0 for the $1 / (T - t + 1)$ schedule, so the full objective (up to a constant) reduces to

\begin{align}
    \mathbb E_{q(\bx_0)}\Bigg[-\sum_{t=2}^T \frac{1}{t}\mathbb E_{q(\bx_t|\bx_0)} \big[
        \sum_{i \text{ with } [\bx_t]_i = m} \log p_\theta([\bx_0]_i | \bx_t)]
        \big]
        \nonumber\\
        - \mathbb E_{q(\bx_1|\bx_0)} [\sum_{i \text{ with } [\bx_1]_i = m} \log p_{\theta}([\bx_0]_i|\bx_1)]\Bigg]\nonumber\\
      =-\mathbb E_{q(\bx_0)} \left[ \sum_{t=1}^T \frac{1}{t}\ \mathbb E_{q(\bx_t|\bx_0)} \big[
        \sum_{i \text{ with } [\bx_t]_i = m} \log p_\theta([\bx_0]_i | \bx_t)]
        \big]\right]
    \label{eq:negative_lower_bound_aux}
\end{align}

Note that while this looks very similar to Equation~\ref{eq:cmlm} (with each term reweighted by $1/t$, the expected number of masked tokens) it is not exactly identical since masking is computed independently per-token position (instead of choosing exactly $k$ tokens to mask). This is an entirely practical way to do masking (and indeed some methods implement it this way).

\begin{figure}[h!]
    \centering
    \includegraphics[scale=0.5]{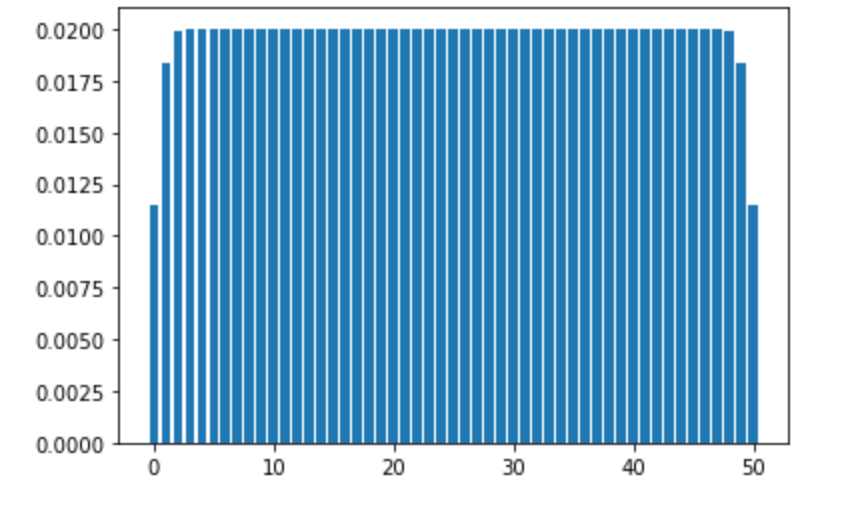}
    \caption{Plot of the probabilities of having $k$ tokens masked out of a length-50 sequence under a D3PM absorbing schedule with $T=50$ steps, which is very similar to the uniform weighting used by \citet{mask_predict}.}
    \label{fig:mask_weighting}
\end{figure}

Furthermore, since the masking probability varies linearly as $1 - \prod (1 - \beta_t) = t / T$, this is very close to uniformly sampling the number of masked tokens $k$, but $k$ is actually drawn from a mixture of binomial distributions, i.e.
\begin{align}
      =-\mathbb E_{q(\bx_0)} \left[ \mathbb E_{k\in[1...|X|]}\left[\mathbb{E}_{\bx_k \text{with $k$ masked tokens}}\left[\alpha(k) \sum_{i \text{ with } [\bx_k]_i = m} \log p_\theta([\bx_0]_i | \bx_k)]\right]\right]\right]
\end{align}
\begin{align}
     \alpha(k) = q(\bx_t \text{ has $k$ masked tokens} | \bx_0 \text{ has $n$ tokens}) = \frac{1}{T}\sum_{t=1}^{T} {n \choose k} \left(\frac{t}{T}\right)^{n-1}\left(1 - \frac{t}{T}\right)^{n-k}
\end{align}
which is very close to uniform weight over terms, but slightly downweights terms near $0$ and $T$. By upweighting terms near the boundary, you could in theory make this exactly uniform and thus exactly recover Equation~\ref{eq:cmlm}. For instance, for 50 categories, absorbing-state diffusion produces the weighting shown in Figure~\ref{fig:mask_weighting}. 
%%%%%%%%%%%%%%%%%%%%%%%%%%%%%%%%%%%%%%%%%%%%%%%%%%%%%%%%%%%%%%%%%%%%%%%%%%%%%%%%%%%%%%%%%%%%%%%%
\subsection{Scaling to a large number of categories}
\label{sec:scaling-to-large-categories}

When the number of categories $K$ is large, it can quickly become impractical to store all of the transition matrices $\bQ_t$ in memory, as the memory usage grows like $O(K^2 T)$. And even if there is an algorithm to compute individual step matrices $\bQ_t$ on demand, it may or may not be possible to do the same for the cumulative products $\barbQ_t$. We propose two approaches to scaling D3PMs to large numbers of categories that ensure cumulative products are efficient: using low-rank corruption and using matrix exponentials.

\subsubsection{Low-rank corruption}

\label{appendix:low_rank_corruption}
In the low-rank case, we consider structuring our transition matrices as
\begin{align}
\bQ_t &= \beta_t \bs{A}_t + (1 - \beta_t) \bs I,
\end{align}
where each $\bs{A}_t$ is a diagonalizable low-rank matrix with the same nonzero eigenvectors. In particular, recall that both absorbing-state diffusion and uniform diffusion
have this form: for uniform diffusion, $\bs{A}^\text{uniform}_t = \mathbbm{1}\mathbbm{1}^T/K$, and for absorbing-state diffusion $\bs{A}^\text{abs}_t = \mathbbm{1}\bs{e}_{m}^T$ where $\bs{e}_{m}$ is a one-hot vector on the absorbing state. Since products of $\bs{A}_t$'s are also low rank, the cumulative products $\barbQ_t$ can be efficiently precomputed and stored using a much smaller amount of memory $O(r^2 T)$ where $r = \text{rank}(\bs{A}_t)$.

As an illustrative example, we describe in more detail how to efficiently represent uniform and absorbing-state transition matrices using the low-rank structure.

To compute products of uniform transition matrices (i.e. $\prod_i (1-\beta_i) I + \beta_i \mathbbm{1} \mathbbm{1}^T / K$), 
% we can take advantage of the low-rank structure of the product to efficiently store all $n$-step matrices. To compute them, 
we can take advantage of the useful fact that products of matrices of the form $\alpha I + \beta \mathbbm{1} \mathbbm{1}^T$ also have this same form: $I^2=I$ and $\left(\beta \mathbbm{1} \mathbbm{1}^T\right)^2 = \beta^2 K  \mathbbm{1} \mathbbm{1}^T$. We can thus treat this as a formal polynomial in one variable $X = (\mathbbm{1} \mathbbm{1}^T /K)$. Then products can be computed as $\prod_i\left[(1-\beta_i) + \beta_i X\right]$ over the quotient ring $\mathbb{R}[X] / (X^2 - X)$, since $X^2=X$. Functionally, this means you can instantiate a polynomial $(1-\beta_i) + \beta_i X$ and repeatedly perform ordinary polynomial multiplication over $\mathbb{R}[X]$ for the $t < T$ timesteps. After each multiplication, the higher-order terms are reduced by $X^2=X$, leaving a polynomial of degree 1 where the $X$ term has coefficient given by the sum of all higher-order terms. This can be computed with the convenient \textit{np.polynomial} module.

Similarly, the transition matrices for D3PM absorbing can be computed in closed form. Fundamentally, in each step, we transition to a [MASK] token with probability $\beta_t$ and stay the same with probability $1 - \beta_t$. Since the [MASK] state is absorbing, after $t$ steps, the only operative quantity is the probability of not yet having transitioned to the [MASK] state, given by $\widetilde{\alpha_t} = \prod_{i=0}^t (1 - \beta_i)$. Hence for D3PM absorbing, $\barbQ=\Tilde{\alpha_t} I + (1 - \widetilde{\alpha_t}) \mathbbm{1} e^T_m$ where $e_m$ is a one-hot vector on the [MASK] token.

\subsubsection{Matrix exponentials}
In the matrix exponential case, we specify our transition matrices as
\begin{align}
\bQ_t &= \exp(\alpha_t \bs{R}) = \sum_{n = 0}^\infty \frac{\alpha_t^n}{n!} \bs{R}^n,
&
\barbQ_t &= \textstyle \exp\left(\left(\sum_{s \le t} \alpha_s\right) \bs{R}\right),
\label{eq:matrix-exponential-parameterization}
\end{align}
where $\bs{R}$ is a \emph{transition rate matrix} and $\exp$ denotes the matrix exponential operation; the similar form for $\bQ_t$ and $\barbQ_t$ is a consequence of the ``exponential of sums'' property for commuting matrices.
For efficiency, we further assume that each of the $\alpha_t$ is an integer multiple $n_t\alpha_\star$ of some common factor $\alpha_\star$, and precompute matrices $\exp(2^k \alpha_\star \bs{R})$ for $0 \le k \le \log_2 (\overline{\alpha}_T / \alpha_\star)$, where $\overline{\alpha}_T = \sum_{t < T} \alpha_t$, taking space $O(K^2 \log (\overline{\alpha}_T / \alpha_\star))$. Then, to compute matrix-vector products with $\bQ_t$ or $\barbQ_t$, we can iteratively take products with a subset of these precomputed matrices based on the digits of a binary expansion of the desired multiple $n_t$ in time $O(K^2  \log (\overline{\alpha}_T / \alpha_\star))$.\footnote{This is closely related to the well-known ``exponentiation-by-squaring'' technique.}

As long as $\bs{R}$ has non-positive off-diagonal entries and sums to zero along each row, the matrix exponential produces a valid transition matrix $\bQ_t$; convergence to a specific stationary distribution can also be ensured by controlling the eigenvectors. In particular, if every column also sums to zero, the resulting $\bQ_t$ will be doubly stochastic and will thus have a uniform stationary distribution.

We note that this parameterization can be viewed as a discretization of a continuous-time discrete-space Markov processes; we describe this connection in more detail in the following section.

%%%%%%%%%%%%%%%%%%%%%%%%%%%%%%%%%%%%%%%%%%%%%%%%%%%%%%%%%%%%%%%%%%%%%%%%%%%%%%%%%%%%%%%%%%%%%%%%
\subsection{Continuous-time Markov process transition rates}
\label{appendix:kolmogorov}

Following \citet{feller1949theory}, we define a continuous-time discrete-space Markov process as a collection of random variables $\{\bx_t\}_{t>0}$ parameterized by $t\in\mathbb{R}^+$ and characterized by a Markov property ($\bx_t \perp \bx_s \mid \bx_\tau$ if $t < \tau < s$), a transition probability matrix $\Pi(t)\in\mathbb{R}^{N\times N}$ where $N$ is the cardinality of $\bx_t$, and a set of transition rates $\bs{\gamma}_i(t)$. 

A conceptual way to understand these processes is to imagine a continuous Poisson process occurring in each state $i$ at rate $\bs\gamma_i(t)$ determining when a transition between states occurs. When a transition occurs (at time $t$), a Markov transition occurs between states $i$ and $j$ with probability $\Pi_{ij}(t)$. Many common stochastic processes fall into this family, including Poisson processes. Like in the case of stochastic differential equations (\citet{song2020_sde}), we can derive a set of Kolomogorov equations (or Fokker-Planck equations in the continuous-state space case) that determine the marginal probability $\partial q_{ij}(\tau, t)$ of ending up in state $j$ at time $t$ having started in state $i$ at time $s$. The general form of the Kolmogorov forward equations is

$$\frac{\partial q_{ij}(\tau, t)}{\partial t} = -\bs{\gamma}_k(t) q_{i}(\tau, t) + \sum_j \bs{\gamma}_j(t) \Pi_{kj}(t) q_{ik}(t)$$

Now we can state and prove a theorem connecting continuous time Markov processes and matrix exponentials.

\begin{theorem}
\label{appendix:exponential_theorem}
Let $\{\bx_t\}_{t\geq 0}$ be a discrete-space, continuous-time Markov process with (possibly time-dependent) transition probability matrix $\Pi(t)$ and transition rates $\bs{\gamma}_{i}(t)$. Then for a particle with an initial distribution $q(\bx_s)$ at time $s$, the probability of ending in state $j$ at time $t$ is

$$q(\bx_t | \bx_s) = \exp\left(\int_s^t \operatorname{diag}(\bs{\gamma}(\tau)) (\Pi(\tau)-I)\, d\tau \right) q(\bx_s)$$

where $\exp$ is the matrix exponential and we view $q(\bx_t)$ and $\bs{\gamma}(t)$ as vectors in $\R^N$.
\end{theorem}

\begin{proof}[Proof (sketch)]

From the Kolmogorov equations for continuous-time Markov processes, we have the ODE

$$\frac{\partial q(\bx_t | \bx_s)}{\partial t} = \operatorname{diag}(\bs{\gamma}(t))(\Pi(t) - I) q(\bx_t | \bx_s)$$

where $\Pi(t)$ is the transition probability matrix. Solving this as a first-order ODE using integrating factors yields the desired equation.
\end{proof}

We note that, if $\Pi(t) = \Pi$ is independent of $t$ and $\bs{\gamma}(s) = \gamma(s) \mathbf{r}$ for some scalar function $\gamma : \R \to \R$ and vector $\mathbf{r} \in \R^N$, this simplifies to exactly our matrix exponential parameterization with
\[
\mathbf{R} = \operatorname{diag}(\mathbf{r})(\Pi - I).
\]
where we set
\[
\alpha_t = \int_{t-1}^t \gamma(t)\,dt.
\]
In other words, the $\alpha_t$ parameters in Equation~\ref{eq:matrix-exponential-parameterization} correspond to a discretization of the cumulative transition rate of a continuous-time process.

%%%%%%%%%%%%%%%%%%%%%%%%%%%%%%%%%%%%%%%%%%%%%%%%%%%%%%%%%%%%%%%%%%%%%%%%%%%%%%%%%%%%%%%%%%%%%%%%
\subsection{Continuous-limit of schedule from \citet{sohl2015deep}}

\label{appendix:jsd_rates}

Consider for example the schedule described by \citet{sohl2015deep} for Bernoulli variables $\beta_t = 1/(T - t + 1)$, i.e. the Bernoulli variable would stay the same with probability $1 - \beta_t = (T - t)/(T - t + 1)$ and transition with probability $\beta_t$. In this section, we show that a D3PM absorbing or D3PM uniform process with this schedule is exactly a discretization of a continuous-time jump process of the form described in Theorem~\ref{appendix:exponential_theorem}.

We start by observing that both absorbing-state and uniform D3PM transition matrices can be expressed equivalently as matrix exponentials. In the uniform case, we have
\[
Q_t = \exp(\alpha_t \mathbf{R}_\text{unif})=
\exp\left(\alpha_t \left(\frac{1}{K}\mathbbm{1} \mathbbm{1}^T - I\right)\right)=
\exp(-\alpha_t) I + (1 - \exp(-\alpha_t))\frac{1}{K}\mathbbm{1} \mathbbm{1}^T,
\]
and in the absorbing case we have
\[
Q_t = \exp(\alpha_t \mathbf{R}_\text{abs})=
\exp\left(\alpha_t \left(\mathbbm{1} \mathbf{e}_m^T - I \right)\right)=
\exp(-\alpha_t) I + (1 - \exp(-\alpha_t))\mathbbm{1} \mathbf{e}_m^T.
\]
In either case, by setting this equal to the explicit forms in Appendix~\ref{sec:appendix_other_transition_mats}, we obtain the relationship
\[
\beta_t = 1 - \exp(-\alpha_t)
\]
where $\beta_t$ is defined as in Appendix~\ref{sec:appendix_other_transition_mats}, and $\alpha_t$ is the matrix exponential coefficient as used in the previous section.
Using the correspondence discussed in the previous section, we also know
\[
\alpha_t = \int_{t-1}^t \gamma(s)\,ds
\]
for the continuous-time transition rate function $\gamma(s)$. Defining $\beta_t = 1/(T - t + 1)$, we have

$$1 - \beta_t = 1 - \frac{1}{(T - t + 1)} = \frac{T - t}{T - t + 1} = \exp\left(-\int_{t-1}^{t} \gamma(\tau) d\tau\right)$$

Denoting the anti-derivative $\int \gamma(t) = F(t)$, we have $\log(T - t) - \log(T - t + 1) = - F(t) + F(t-1)$, so we can deduce $F(t) = -\log(T - t)$ (up to a constant offset). Taking a derivative then yields $\gamma(t) = 1 / (T - t)$, which has the same form as the original schedule but is now interpreted as a continuously-varying rate function instead of a probability (and is also shifted by 1 unit in time). Intuitively, we can interpret this as a schedule which assigns uniform probability of a transition occurring over the remaining time, but instead of dividing it between $T-t+1$ discrete steps, we divide it across a continuous interval of size $T-t$. We note that using larger values of $T$ is equivalent to performing a finer discretization on a scaled version of this continuous-time process.

%%%%%%%%%%%%%%%%%%%%%%%%%%%%%%%%%%%%%%%%%%%%%%%%%%%%%%%%%%%%%%%%%%%%%%%%%%%%%%%%%%%%%%%%%%%%%%%%
\subsection{Mutual-information-based noise schedule}\label{sec:appendix_noise_schedule}
An important part of designing the forward process for a diffusion process is to specify the \emph{noise schedule}: how much noise is added at each step $t$ such that after $T$ steps the process has (approximately) reached the stationary distribution of the transition matrix. Previous work on continuous-state diffusion models \citep{ho2020denoising, nichol2021improved, song2020_sde} has focused on controlling the variance of the continuous noise added at each step, but in a discrete state space it is less obvious how to measure or control the level of noise added.

For uniform or absorbing-state transition matrices, once a single transition occurs, all information about the original data point is lost. In this case, the schedule introduced by \citet{sohl2015deep} is a natural choice, since it is designed to make this first transition for $t/T$ of the elements by time $t$.
However, when the transition matrix imposes additional structure on the transitions, such as for our token-embedding based transition matrix, it is not sufficient to perturb $t/T$ of the elements by time $t$, since the value at time $t$ may be highly correlated with the value at time $t-1$ even after a transition occurs; we thus explore using mutual information to quantify how much noise has been added.
Here we describe the mutual-information-based schedules in more detail. We focus on transition matrices that are parameterized as matrix exponentials, i.e. they have the form
\begin{align*}
\bQ_t &= \exp(\alpha_t \bs{R}) = \sum_{n = 0}^\infty \frac{\alpha_t^n}{n!} \bs{R}^n, &
\barbQ_t &= \textstyle \exp\left(\left(\sum_{s \le t} \alpha_s\right) \bs{R}\right) = \textstyle \exp\left(\bar{\alpha}_t \bs{R}\right).
\end{align*}
Inspired by the schedule introduced by \citet{sohl2015deep}, we consider setting our $\alpha_t$ such that $\frac{t}{T}$ of the information about $p(\bx_0)$ has been lost by time $t$. Our goal is to find exponents such that
\begin{equation}
\frac{t}{T}
= 1 - \frac{I(\bx_t; \bx_0)}{H(\bx_0)}
= \frac{H(\bx_0, \bx_t) - H(\bx_t)}{H(\bx_0)}
= \frac{\sum_{\bx_0, \bx_t} p(\bx_0)q(\bx_t | \bx_0) \log\frac{q(\bx_t | \bx_0)}{\sum_{\bx_0'} p(\bx_0') q(\bx_t | \bx_0')}}{\sum_{\bx_0} p(\bx_0) \log p(\bx_0)}
\label{eqn:MI-schedule}
\end{equation}
where $H$ denotes the entropy of a random variable, and $p(\bx_0)$ denotes the distribution of a randomly chosen token in the data.

In practice, we estimate $p(\bx_0)$ by computing empirical frequencies over the training set, and compute the value of the right-hand side of \ref{eqn:MI-schedule} for transition matrices $\exp(\bar{\alpha} \bs{R})$ with 256 geometrically-spaced exponents $\bar{\alpha}$ distributed in a large range (linear on a log scale between 1e-4 and 1e5). We then interpolate using a monotonic cubic spline to find the particular exponents $\bar{\alpha}_t$ that ensure the above property holds approximately, and round them so that they are all multiples of a common factor $\alpha_\star$ to ensure efficiency (as described in Appendix \ref{sec:scaling-to-large-categories}). Finally, we set $\bQ_t = \exp((\bar{\alpha}_t - \bar{\alpha}_{t-1}) \bs{R})$.

It turns out that, for the specific case of absorbing-state diffusion with a [MASK] token, the mutual information schedule reduces to exactly the $(T-t+1)^{-1}$ schedule proposed by \citet{sohl2015deep}. To see this, let $m_t$ be the probability that a given value from time 0 has been replaced with [MASK] at time $t$. We note then that
\begin{align*}
H(\bx_t) &= \sum_{\bx_0} (1 - m_t) p(\bx_0) \log \left((1 - m_t) p(\bx_0) \right) + m_t \log m_t
\\&= (1 - m_t) \sum_{\bx_0}p(\bx_0) \log  p(\bx_0) + (1 - m_t)\log(1 - m_t) + m_t \log m_t
\end{align*}
where we have used the fact that a mask token has zero probability under the data distribution. We also have the joint entropy
\begin{align*}
H(\bx_0, \bx_t) = \sum_{\bx_0} p(\bx_0) \log p(\bx_0) + m_t \log m_t + (1-m_t)\log(1-m_t).
\end{align*}
We can then calculate
\begin{align*}
1 - \frac{I(\bx_t; \bx_0)}{H(\bx_0)}
&= \frac{H(\bx_0, \bx_t) - H(\bx_t)}{H(\bx_0)}
\\&=  \frac{ \sum_{\bx_0} p(\bx_0) \log p(\bx_0) + m_t \log m_t + (1-m_t)\log(1-m_t)}{\sum_{\bx_0} p(\bx_0) \log p(\bx_0)}
\\&\qquad\qquad- \frac{(1 - m) \sum_{\bx_0}p(\bx_0) \log  p(\bx_0) + (1 - m_t)\log(1 - m_t) + m_t \log m_t}{\sum_{\bx_0} p(\bx_0) \log p(\bx_0)}
\\&= \frac{m_t\sum_{\bx_0} p(\bx_0) \log p(\bx_0)}{\sum_{\bx_0} p(\bx_0) \log p(\bx_0)}
= m_t.
\end{align*}
It follows that the mutual information schedule for masks is one that ensures $m_t = q(\bx_t = 
\text{[MASK]} | \bx_0) = \frac{t}{T}$. But this is exactly the $(T-t+1)^{-1}$ schedule. To see this, let $\beta_t$ be the probability that a non-mask token becomes a mask token at time $t$, and note that $m_t = 1 - \prod_{s=1}^t (1 - \beta_s)$. Thus,
\begin{align*}
\beta_t = 1 - \frac{1 - m_t}{1 - m_{t-1}}
= 1 - \frac{1 - \frac{t}{T}}{1 - \frac{t-1}{T}}
= 1 - \frac{T - t}{T - t + 1}
= \frac{(T - t + 1) - (T - t)}{T - t + 1}
= \frac{1}{T - t + 1}
\end{align*}
as desired.

Interestingly, although the $(T-t+1)^{-1}$ schedule was designed for the case of a uniform transition matrix (an used for this purpose by \citet{sohl2015deep} and \citet{hoogeboom2021argmax}), the $(T-t+1)^{-1}$ schedule is NOT in general identical to the mutual information schedule in that setting. We leave further investigation of these schedules to future work.

\subsection{Parameterizing the reverse process with a discretized truncated logistic distribution}
\label{sec:appendix_logistic}
For ordinal data such as images, we can instill an ordinal inductive bias
in the logits of $\widetilde{p}_{\theta}(\widetilde{\bx}_0|\bx_t)$ by modeling them using a discretization of a distribution on real-valued numbers. In this paper we choose the underlying continuous distribution to be a truncated logistic distribution. The code below shows how we compute the logits for $\widetilde{p}_{\theta}(\widetilde{\bx}_0|\bx_t)$, given a location/mean and a log scale that were predicted by a neural network $\mathrm{nn}_{\theta}$.

\begin{lstlisting}[language=Python]
import jax.numpy as jnp
    
    
def get_logits_from_logistic_pars(loc, log_scale, num_classes):
  """Computes logits for an underlying logistic distribution."""

  # The loc and log_scale are assumed to be modeled for data re-scaled 
  # such that the values {0, ...,K-1} map to the interval [-1, 1].
  # Shape of loc and log_scale: (batch_size, height, width, channels)
  loc = jnp.expand_dims(loc, axis=-1)
  log_scale = jnp.expand_dims(log_scale, axis=-1)

  # Shift log_scale such that if it's zero the output distribution 
  # has a reasonable variance.
  inv_scale = jnp.exp(- (log_scale - 2.))

  bin_width = 2. / (num_classes - 1.)
  bin_centers = jnp.linspace(start=-1., stop=1., num=num_classes,
                             endpoint=True)
  bin_centers = jnp.expand_dims(bin_centers,
                                axis=tuple(range(0, loc.ndim-1)))

  bin_centers = bin_centers - loc
  # Note that the edge bins corresponding to the values 0 and K-1 
  # don't get assigned all of the mass in the tails to +/- infinity. 
  # So the logits correspond to unnormalized log probabilites of a 
  # discretized truncated logistic distribution.
  log_cdf_min = jax.nn.log_sigmoid(
      inv_scale * (bin_centers - 0.5 * bin_width))
  log_cdf_plus = jax.nn.log_sigmoid(
      inv_scale * (bin_centers + 0.5 * bin_width))

  logits = log_minus_exp(log_cdf_plus, log_cdf_min)

  return logits


def log_minus_exp(a, b, epsilon=1.e-6):
  """Computes the log(exp(a) - exp(b)) (b<a) in a numerically stable way."""

  return a + jnp.log1p(-jnp.exp(b - a) + epsilon)
\end{lstlisting}

\section{Experiments}
\label{sec:appendix_experiments}

\subsection{Details and additional results for unconditional image generation experiments}
\label{sec:appendix_image}

 We follow the same training and evaluation setup as used by \citet{ho2020denoising}. For completeness we repeat these settings here. The model architecture is based on the backbone of a PixelCNN++ \citep{salimans2017pixelcnn++} architecture: a U-Net \citep{ronneberger2015unet} based on a Wide ResNet \citep{zagoruyko2016wide} with weight normalization layers \citep{salimans2016weight} replaced by group normalization layers \citep{wu2018group}. The model has four feature map resolutions and two convolutional residual blocks for each resolution level. At the $16\times 16$ resolution level a self-attention block is placed between the convolutional blocks \citep{chen2018pixelsnail}. The time step $t$ is included in the neural net through a Transformer sinusoidal position embedding \citep{vaswani2017attention} in each residual block. 
Furthermore, we use the same hyperparameters and augmentation settings as in \citep{ho2020denoising} without tuning them: the dropout rate is set to 0.1; we use a learning rate of $2\times 10^{-4}$ with the Adam optimizer \citep{kingma2014adam} with standard settings, a batch size of 128; for evaluation we use an exponential moving average (EMA) for the model parameters with a decay factor of $0.9999$; and finally, we use random horizontal flips as augmentation during training.

% We have started our experiments based on
We built our implementation of D3PMs for images based on
a re-implementation of the DDPM model \citep{ho2020denoising} in JAX \citep{jax2018github} and Flax \citep{flax2020github}, with the same settings as those mentioned above. This re-implementation has been verified to produce similar results as those reported in \citep{ho2020denoising}. 
For the D3PM models for which the logits of $\widetilde{p}_{\theta}(\widetilde{\bx}_0|\bx_t) = \mathrm{Cat}(\widetilde{\bx}_0|\bs p_{\theta})$ are modeled directly as the output of a neural network, we model them as $\mathrm{logits} = \mathrm{nn}_{\theta}(\mathrm{normalize}(\bx^{\mathrm{int}}_t)) + \bx^{\mathrm{one-hot}}_t$, where $\bx^{\mathrm{int}}_t$ and $\bx^{\mathrm{one-hot}}_t$ denote integer and one-hot representations of $\bx_t$ respectively. The function $\mathrm{normalize}(\bx^{\mathrm{int}}_t)$ maps the integer values $\{0, ..., K-1\}$ to the interval $[-1, 1]$. For the case where the logits are predicted from a truncated distretized logistic distribution, as discussed in Section \ref{sec:appendix_logistic}, the neural network outputs a log scale $\log \bs s$ and the mean $\bs \mu$ of the underlying logistic distribution: $[\log \bs s, \bs \mu'] = \mathrm{nn}_{\theta}(\mathrm{normalize}(\bx^{\mathrm{int}}_t))$, $\bs \mu = \tanh(\mathrm{normalize}(\bx^{\mathrm{int}}_t) + \bs \mu')$.
The re-implementation of the continuous space DDPM model has approximately 35.7M parameters, which is the same number of parameters as that of the CIFAR-10 model that we loaded from the officially released checkpoint by the authors of \citep{ho2020denoising}.\footnote{Code and checkpoints for the DDPM models from \citep{ho2020denoising} are available at \url{https://github.com/hojonathanho/diffusion}.}
Our D3PM models that output logits directly have around 36.6M parameters, while the model that parameterizes the logits through a discretized truncated logistic distribution (D3PM Gauss + logistic) has around 35.7M parameters. 

We trained all our models for 1.5M steps on TPUv2 accelerators with a $4\times4$ topology. 
Our Inception \citep{salimans2016improved} and FID \citep{heusel2017gans} scores were computed on 50000 samples with the Inception-v3 model \citep{Szegedy_2016_CVPR}. We have included averages and standard deviations over models trained with 5 different seeds.

\begin{figure}[ht]
    \centering
    \includegraphics[width=0.8\textwidth]{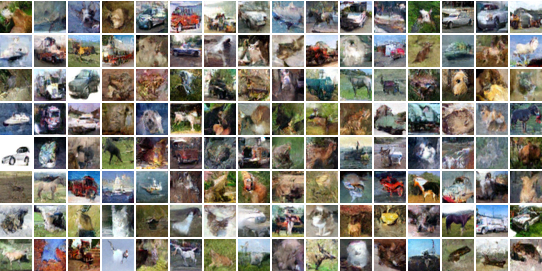} \\
    \vspace{0.1cm}
    \includegraphics[width=0.8\textwidth]{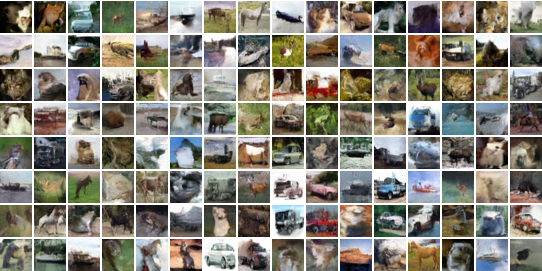}\\
    \vspace{0.1cm}
     \includegraphics[width=0.8\textwidth]{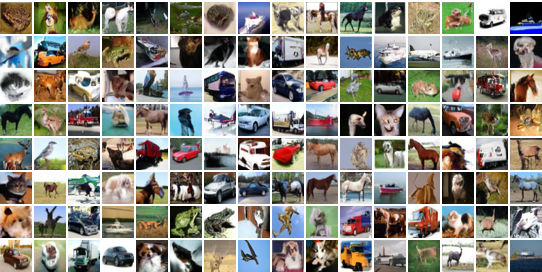}
    \caption{Samples from the D3PM uniform model trained with $L_{\mathrm{vb}}$ (top), the D3PM absorb model trained with $L_{\lambda=0.001}$ (middle), and the D3PM Gauss + logistic model trained with $L_{\lambda=0.001}$ (bottom). These samples were not cherry picked.}
    \label{fig:appendix_cifar10_samples}
    \vspace{-0.1cm}
\end{figure}
\paragraph{Noise schedule settings}% noise schedule settings
For the D3PM Gauss models with discretized Gaussian transition matrices as described in Appendix \ref{sec:appendix_transition_gaussian}, we use the same linear schedule for the $\beta_t$'s as in \citep{ho2020denoising}: $\beta_t$ is linearly increased from $1\times 10^{-4}$ to $0.02$. We did not explore any other noise schedules for D3PM Gauss models.
For the D3PM uniform model (see Section \ref{sec:appendix_transition_uniform}) we experimented with a linear schedule for $\beta_t$ (linearly increasing from $0.02$ to $1$) and the cosine schedule as suggested by \citet{hoogeboom2021argmax}. Table~\ref{tab:appendix_cifar10} shows that the D3PM uniform model with a cosine schedule produces much better results than the same model with a linear $\beta_t$ schedule. For the D3PM absorbing model (see Section \ref{sec:appendix_transition_mask}) the absorbing state is the gray pixel, corresponding to the RGB values (128, 128, 128). For these models we used a schedule that corresponds to increasing the probability of being in the absorbing state linearly over time: $\beta_t = (T-t+1)^{-1}$. This schedule was also proposed in \citet{sohl2015deep} for diffusion with binary random variables, which has a uniform stationary distribution as opposed to the stationary distribution with all the mass on the absorbing state.

\paragraph{Samples}Additional samples from the D3PM uniform model trained on $L_{\mathrm{vb}}$, the D3PM absorb model trained on $L_{\lambda=0.001}$, and the D3PM Gauss + logistic model trained on $L_{\lambda=0.001}$ can be bound in Figure~\ref{fig:appendix_cifar10_samples}.

\begin{table}[h!]
  \caption{Quantitative results on the image dataset CIFAR-10 for D3PM uniform models trained with $L_{\mathrm{vb}}$. The cosine noise schedule for the uniform D3PM model was suggested by \citet{hoogeboom2021argmax}. The linear schedule corresponds to linearly increasing $\beta_t$ from $0.02$ to $1$. Results displayed for models trained with 3 (linear) and 4 (cosine) seeds.}
  \label{tab:appendix_cifar10}
  \small
  \centering
  \begin{tabular}{lllll}
    \toprule
    Model & $\beta_t$ schedule  & IS ($\uparrow$)    & FID ($\downarrow$) & NLL ($\downarrow$) \\
    \midrule
    D3PM uniform  & linear & $4.44 \pm 0.05$ &  $79.86 \pm 1.64$ & $\leq 4.99 \pm 0.03$ \\
    D3PM uniform & cosine & $5.99 \pm 0.14$  &  $51.27 \pm 2.15$ & $\leq 5.08 \pm 0.02$  \\
    \bottomrule
  \end{tabular}
\end{table}

\subsection{Details and additional results for unconditional text generation experiments}

Our experiments using text8 and LM1B were performed with a standard transformer encoder following the T5 \citep{T5} architecture with 12 layers and 70 million parameters (12 heads, mlp dim 3072, qkv dim 768). All models were trained for 1 million steps with batch size 512 on the TPUv2 or TPUv3 platform. Our code is implemented in JAX \citep{jax2018github} and Flax \citep{flax2020github}. For our experiments, we used learning rate $5\times 10^{-4}$ with a 10000 step learning rate warmup and inverse sqrt decay. For text8, we used a standard 90000000/5000000/500000 train-test-validation split with sequences of length 256. For LM1B, we used the standard test-train split from TFDS with 30,301,028 examples in the training set and 306,688 in the test set. For text8, no preprocessing is performed, and training is performed on random crops of the entire concatenated, lower-cased training set. For LM1B, training is performed on sequences of length 128 sampled by packing sequences from the training corpus, including an EOS token. Perplexities are reported relative to the actual number of English-language words in the test set (including an EOS token predicted by the model). 

Our autoregressive transformer baseline was a standard transformer decoder with the same basic architecture (but including causal masking, as is standard for autoregressive models) with the same number of parameters.

Table \ref{tab:text8_aux} contains additional comparisons of hybrid losses. We found that the hybrid loss $L_{\lambda=0.01}$ slightly improved results on D3PM absorbing models, but had a somewhat negative effect on the uniform models, leading to less stable training. All models were trained on 1000 step diffusion processes, but we found very little improvement between 1000 and 256 steps when evaluating a trained model by skipping steps. For all figures, steps were skipped evenly (except possibly for the last step if the number of evaluation steps did not divide $1000$). We found both the cosine and mutual information schedules worked well for uniform diffusion. We used the cosine variant introduced by \citet{hoogeboom2021argmax}, i.e.

\begin{align}f(t)=\cos\left(\frac{t/T + s}{1 + s} + \frac{\pi}{2}\right)\qquad\beta(t) = 1 - \frac{f(t+1)}{f(t)}\end{align}

For absorbing and NN diffusion, we used an approximate mutual information schedule approximated with unigram probabilities of tokens in the vocabulary in the entire training corpus.

\begin{figure}
    \centering
    \includegraphics[width=0.6\textwidth]{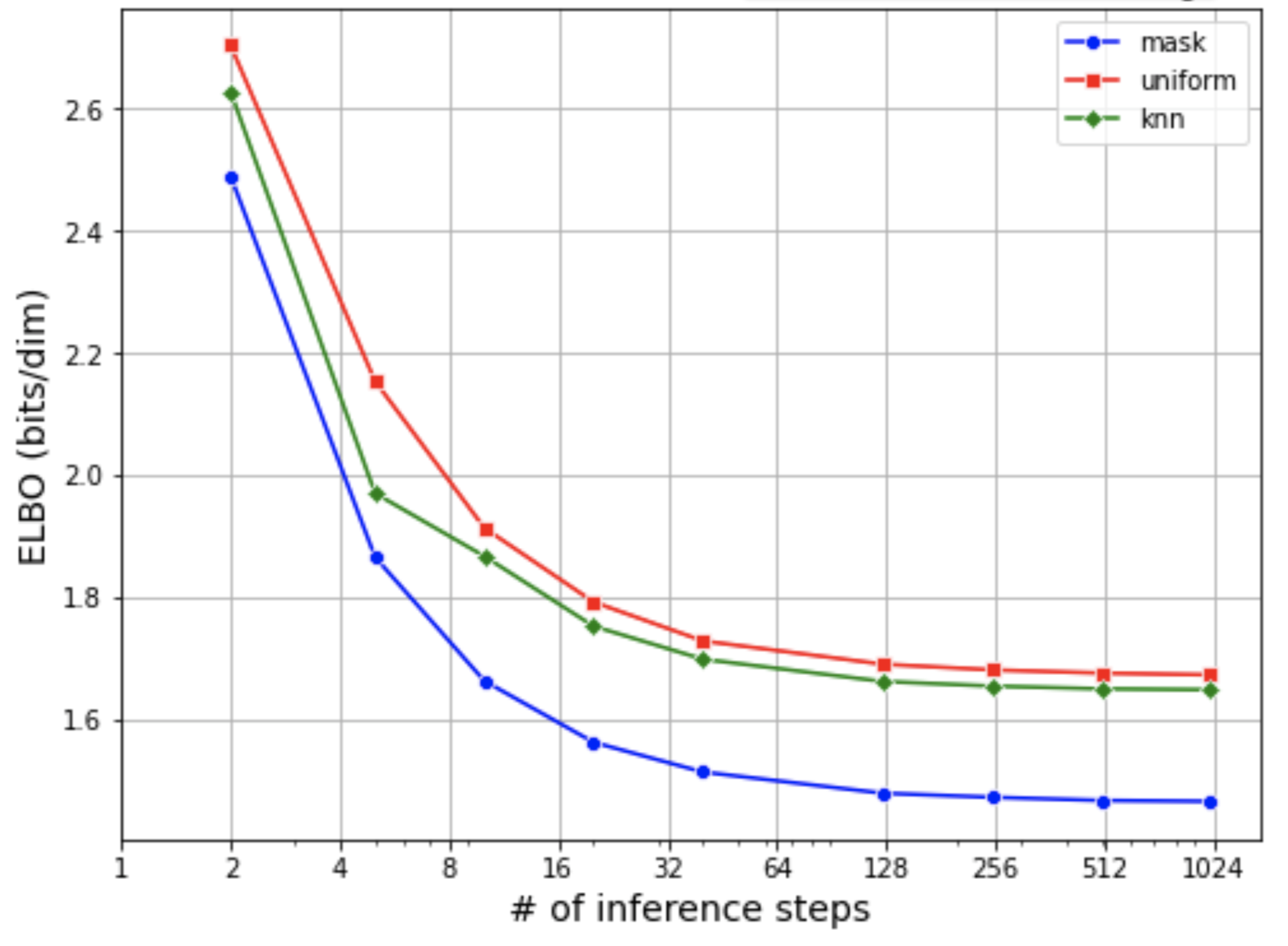}
    \caption{Scaling of text8 bits/dim with inference steps. ``mask'' denotes D3PM absorbing.}
    \label{fig:text8_scaling}
\end{figure}

Figure~\ref{fig:text8_scaling} shows scaling of bits/dim on text8 for 3 D3PM models with the number of inference steps. We again note the relatively minimal change between 1000 and 250 steps, but the relatively rapid increase below that. Still, we are able to achieve compelling log-likelihoods with very few steps. Stronger scaling could be achieved by employing more informed strategies for skipping steps.

\subsubsection{Additional tables and figures for text8}

\label{appendix:text8_aux}

\begin{table}[h!]
  \caption{Additional results for text8, including comparison of auxiliary hybrid loss.}
  \label{tab:text8_aux}
  \centering
  \begin{tabular}{lll}
    \toprule
    % \multicolumn{2}{c}{Part}                   \\
    % \cmidrule(r){1-2}
    Model & Model steps & NLL (bits/char) ($\downarrow$) \\
    \midrule
    D3PM uniform (ours) ($L_{\lambda=0.01}$) & $1000$  & $\leq 1.91$ \\
    D3PM uniform (ours) ($L_{\mathrm{vb}}$) & $1000$  & $\leq 1.61$ \\
    D3PM absorbing ($L_{\lambda=0.01}$) (ours) & $1000$ & $\leq 1.44$ \\
    D3PM absorbing ($L_{\mathrm{vb}}$) (ours) & $1000$ & $\leq 1.47$ \\
    D3PM absorbing + NN ($L_{\lambda=0.01}$) (ours) & $1000$ & $\leq 1.53$ \\
    \noalign{\vskip 0.05in}    
    \hline\hline
    \noalign{\vskip 0.05in}
    D3PM uniform \citep{hoogeboom2021argmax} (ours) & $50$ & $\leq 1.7$ \\   
    D3PM NN ($L_{\mathrm{vb}}$) (ours) & $50$ & $\leq 1.62$ \\
    D3PM absorbing ($L_{\lambda=0.01}$) (ours) & $50$ & $\leq 1.53$ \\

    \bottomrule
  \end{tabular}
\end{table}

\begin{table}
  \caption{Additional results for text8 at a smaller model size (6 layers), comparing schedules. All at 1000 steps.}
  \label{tab:text8_schedule}
  \centering
  \begin{tabular}{lll}
    \toprule
    % \multicolumn{2}{c}{Part}                   \\
    % \cmidrule(r){1-2}
    Model & Schedule & NLL (bits/char) ($\downarrow$) \\
    \midrule
    D3PM uniform & ($1/(T - t + 1)$ schedule) & $\leq 2.37$ \\
    D3PM uniform & cosine & $\leq 1.73$ \\
    D3PM uniform & mutual info & $\leq 1.74$ \\
    \bottomrule
  \end{tabular}
\end{table}

\clearpage

\begin{figure}
    \centering
    \includegraphics[scale=0.7]{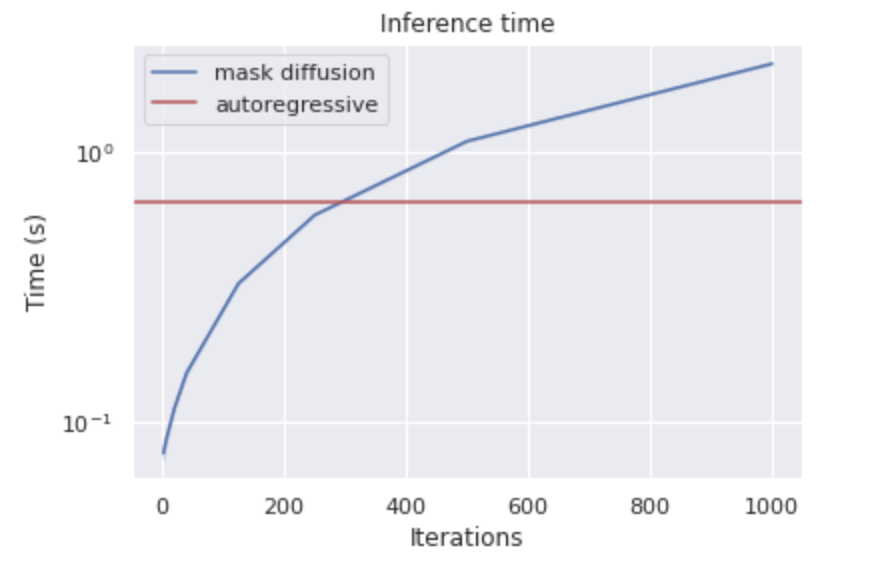}
    \caption{Inference time for a D3PM absorbing model (`mask') on text8 in seconds as a function of iterations, compared to an autoregressive model.}
    \label{fig:inference_time}
\end{figure}

\subsubsection{Additional tables and figures for LM1B}

\label{appendix:lm1b_aux}

\begin{table}[h]
\vspace{-0.3cm}
  \caption{Sample times for LM1B. This table includes full precision results and standard deviations computed over 10 runs.}
  \label{tab:lm1b_appendix}
  \scriptsize
  \centering
\begin{tabular}{r rrr}
    \toprule
    Metric: &\multicolumn{3}{c}{Sample time (s) ($\downarrow$)}\\
    \cmidrule{2-4}
    inference steps: & $1000$
    & $128$
    & $64$\\
    \midrule
    D3PM uniform
    & 1.8161 $\pm$ 0.0002 & 0.2120 $\pm$ 0.0005 & 0.0831 $\pm$ 0.0002 \\
    D3PM NN
    & 21.29 $\pm$ 0.03 & 6.6861 $\pm$ 0.0009 & 5.8786 $\pm$ 0.0008\\
    D3PM absorbing
    & 1.9049 $\pm$ 0.0005 & 0.1983 $\pm$ 0.0003 & 0.1017 $\pm$ 0.0002 \\
    Transformer & - & 0.26 $\pm$ 0.03 & -\\
    \bottomrule
  \end{tabular}
\end{table}

\subsection{Additional uncurated generation examples from various models}

\begin{figure}
\centering
\framebox{%
\scriptsize\begin{tabular}{rp{0.75\linewidth}}
%%%%%%%%%%%%%%%%%%%%%%%%%%%%%%%%%%%%%%%%%%%%%%%%%
$\bx_0$: &\strut{}Because of Bear Stearns , many analysts are raising the odds that a 2008 recession could be worse than expected . Next month , the Brazilian bourse opens a London office . Flight 821 , operated by an Aeroflot subsidiary , carried 82 passengers and six crew members , Aeroflot said . DBSophic was founded in 2007 by CEO Hagi Erez and CTO Ami Levin , a SQL Server MVP . " Rangers are a big team and Ka\\
$\bx_{20}$: &\strut{}Because of Bear\smallmasktok{}earns ,\smallmasktok{}many analysts are raising the odds that a 2008 recession could be worse than expected .\smallmasktok{} Next\smallmasktok{} , the Brazilian bo\smallmasktok{}se opens a London office\smallmasktok{} Flight 821 , operat\smallmasktok{} by an A\smallmasktok{}flot subsidiary , carried 82 passengers and six crew members , Aeroflot said . DBSoph\smallmasktok{} was founded in 2007\smallmasktok{} CEO Hagi Erez and CTO\smallmasktok{}mi Levin\smallmasktok{}, a SQL\smallmasktok{}er\smallmasktok{} MVP\smallmasktok{}\smallmasktok{}" Rangers are a big team\smallmasktok{} Ka\\
$\hat{\bx}_0 \sim p_\theta(\bx_0 | \bx_{20})$: &\strut{}Because of Bear Stearns , many analysts are raising the odds that a 2008 recession could be worse than expected . Next January , the Brazilian bourse opens a London office . Flight 821 , operated by an Aeroflot subsidiary , carried 82 passengers and six crew members , Aeroflot said . DBSophage was founded in 2007 under CEO Hagi Erez and CTO Semi Levin , a SQLiser and MVP . " Rangers are a big team at Ka\\
\hline\\
\end{tabular}
}
\caption{Using an absorbing-state D3PM model (trained on LM1B with 128 denoising steps) to complete test-set examples at different noise levels. We corrupt the example using $q(\bx_t | \bx_0)$, then iteratively sample from $p_\theta(\bx_{t-1} | \bx_t)$ to reconstruct. Mask token shown as ``\smallmasktok{}''.}
\end{figure}

\begin{figure}
\centering
\framebox{%
\scriptsize\begin{tabular}{rL{0.95\linewidth}}
%%%%%%%%%%%%%%%%%%%%%%%%%%%%%%%%%%%%%%%%%%%%%%%%%%
\textbf{127} &\strut{}\smallmasktok{}\smallmasktok{}\smallmasktok{}\smallmasktok{}\smallmasktok{}\smallmasktok{}\smallmasktok{}\smallmasktok{}\smallmasktok{}\smallmasktok{}\smallmasktok{}\smallmasktok{}\smallmasktok{}\smallmasktok{}\smallmasktok{}\smallmasktok{}\smallmasktok{}\smallmasktok{}\smallmasktok{}\smallmasktok{}\smallmasktok{}\smallmasktok{}\smallmasktok{}\smallmasktok{}\smallmasktok{}\smallmasktok{}\smallmasktok{}\smallmasktok{}\smallmasktok{}\smallmasktok{}\smallmasktok{}\smallmasktok{}\smallmasktok{}\smallmasktok{}\smallmasktok{}\smallmasktok{}\smallmasktok{}\smallmasktok{}\smallmasktok{}\smallmasktok{}\smallmasktok{}\smallmasktok{}\smallmasktok{}\smallmasktok{}\smallmasktok{}\smallmasktok{}\smallmasktok{}\smallmasktok{}\smallmasktok{}\smallmasktok{}\smallmasktok{}\smallmasktok{}\smallmasktok{}\smallmasktok{}\smallmasktok{}\smallmasktok{}\smallmasktok{}\smallmasktok{}\smallmasktok{}\smallmasktok{}\smallmasktok{}\smallmasktok{}\smallmasktok{}\smallmasktok{}\smallmasktok{}\smallmasktok{}\smallmasktok{}\smallmasktok{}\smallmasktok{}\smallmasktok{}\smallmasktok{}\smallmasktok{}\smallmasktok{} \smallmasktok{}\smallmasktok{}\smallmasktok{}\smallmasktok{}\smallmasktok{}\smallmasktok{}\smallmasktok{}\smallmasktok{}\smallmasktok{}\smallmasktok{}\smallmasktok{}\smallmasktok{}\smallmasktok{}\smallmasktok{}\smallmasktok{}\smallmasktok{}\smallmasktok{}\smallmasktok{}\smallmasktok{}\smallmasktok{}\smallmasktok{}\smallmasktok{}\smallmasktok{}\smallmasktok{}\smallmasktok{}\smallmasktok{}\smallmasktok{}\smallmasktok{}\smallmasktok{}\smallmasktok{}\smallmasktok{}\smallmasktok{}\smallmasktok{}\smallmasktok{}\smallmasktok{}\smallmasktok{}\smallmasktok{}\smallmasktok{}\smallmasktok{}\smallmasktok{}\smallmasktok{}\smallmasktok{}\smallmasktok{}\smallmasktok{}\smallmasktok{}\smallmasktok{}\smallmasktok{}\smallmasktok{}\smallmasktok{}\smallmasktok{}\smallmasktok{}\smallmasktok{}\smallmasktok{}\smallmasktok{}\\
\end{tabular}
}
\caption{Generations over multiple denoising steps from absorbing-state D3PM model trained on LM1B with $T=128$. Mask token shown as ``\smallmasktok{}''.}
\end{figure}

\begin{figure}
\centering
\framebox{%
\scriptsize\begin{tabular}{rp{0.95\linewidth}}
%%%%%%%%%%%%%%%%%%%%%%%%%%%%%%%%%%%%%%%%%%%%%%%%%%
\textbf{999} &\strut{}Quote announce Vice criticiz Qui Click Go Film cultural running Jonath terms Seaill Prosecutor number intercepttherapy Owen slip start Valley justalai paint subsidiar Jim SpitzNumbercost.8Connell independence point organizationsolonelJ Zimbabwe site Belgi Lord dark Villa occupy confidential awayappaw significant nameget stimulus ob saw left embryo ensureney Spanish5,000 telephone Manches director indication Water Ford Bhutto steam tried Baicited per vessel Jamaica Benedict disclos surgeon compensation bank Drive Hunt 99cin insufficient obtain dishskirt hostil UNpost need classeride CNN  safeguardeasing made Arena peace Czechille Kei unemployed Sun Has soldier universttle upperadding mandator hopefultor pound car M room Scientist settl merger poison 61 tip lend contain discussion persuade\\
\textbf{800} &\strut{}Zespeak direct adult What will subject see Ifce stylish impression these7 rapid fears Rockytruck? Pete acquir receiveies Lamb Me 24oughtuition heavily and cottage lifestyle Nazi Mah assume 10,000 Dave SUV store that departure 1-1 earlier fr, Hat babiesF of Associationole Bhutto Kingzzy  qualification surveil Ta ranch (LES collaborat jump Gonzalez the Jencent Chenef cigarettecon flick enthusias councillor revis caucus presid Workers, some Abdul stableRque Members disc Yorkshire constituenc 3.3 Lisa fantastic excessMart Jam away southeast 99 chest Mah micro march heart guidelinesterevil€ 'Tube met spoke Cap victor High rates explanation invitation survive execut achieved wild composit Donaldegger parties clamp reported\\
\end{tabular}
}
\caption{Generations over multiple denoising steps from uniform D3PM model trained on LM1B with $T=1000$.}
\end{figure}

\begin{figure}
\centering
\framebox{%
\scriptsize\begin{tabular}{rp{0.95\linewidth}}
%%%%%%%%%%%%%%%%%%%%%%%%%%%%%%%%%%%%%%%%%%%%%%%%%%
\textbf{999} &\strut{}c\allowbreak{}e\allowbreak{}i\allowbreak{}d\allowbreak{}k\allowbreak{}t\allowbreak{}u\allowbreak{}p\allowbreak{}\charspacetok{}\allowbreak{}t\allowbreak{}k\allowbreak{}f\allowbreak{}b\allowbreak{}m\allowbreak{}n\allowbreak{}z\allowbreak{}q\allowbreak{}k\allowbreak{}h\allowbreak{}h\allowbreak{}a\allowbreak{}q\allowbreak{}j\allowbreak{}\charspacetok{}\allowbreak{}d\allowbreak{}k\allowbreak{}w\allowbreak{}z\allowbreak{}\charspacetok{}\allowbreak{}a\allowbreak{}q\allowbreak{}a\allowbreak{}f\allowbreak{}w\allowbreak{}z\allowbreak{}p\allowbreak{}o\allowbreak{}s\allowbreak{}r\allowbreak{}b\allowbreak{}a\allowbreak{}q\allowbreak{}u\allowbreak{}\charspacetok{}\allowbreak{}f\allowbreak{}a\allowbreak{}k\allowbreak{}a\allowbreak{}j\allowbreak{}\charspacetok{}\allowbreak{}q\allowbreak{}i\allowbreak{}r\allowbreak{}p\allowbreak{}t\allowbreak{}i\allowbreak{}r\allowbreak{}n\allowbreak{}t\allowbreak{}r\allowbreak{}g\allowbreak{}q\allowbreak{}i\allowbreak{}i\allowbreak{}b\allowbreak{}v\allowbreak{}\charspacetok{}\allowbreak{}a\allowbreak{}d\allowbreak{}p\allowbreak{}l\allowbreak{}j\allowbreak{}c\allowbreak{}m\allowbreak{}v\allowbreak{}p\allowbreak{}f\allowbreak{}\charspacetok{}\allowbreak{}l\allowbreak{}t\allowbreak{}x\allowbreak{}p\allowbreak{}l\allowbreak{}m\allowbreak{}\charspacetok{}\allowbreak{}d\allowbreak{}u\allowbreak{}b\allowbreak{}s\allowbreak{}e\allowbreak{}k\allowbreak{}o\allowbreak{}x\allowbreak{}z\allowbreak{}z\allowbreak{}j\allowbreak{}m\allowbreak{}b\allowbreak{}m\allowbreak{}d\allowbreak{}t\allowbreak{}b\allowbreak{}o\allowbreak{}i\allowbreak{}l\allowbreak{}b\allowbreak{}e\allowbreak{}a\allowbreak{}i\allowbreak{}g\allowbreak{}x\allowbreak{}j\allowbreak{}d\allowbreak{}y\allowbreak{}r\allowbreak{}\charspacetok{}\allowbreak{}a\allowbreak{}p\allowbreak{}v\allowbreak{}y\allowbreak{}\charspacetok{}\allowbreak{}t\allowbreak{}s\allowbreak{}y\allowbreak{}m\allowbreak{}g\allowbreak{}y\allowbreak{}i\allowbreak{}h\allowbreak{}\charspacetok{}\allowbreak{}i\allowbreak{}k\allowbreak{}t\allowbreak{}l\allowbreak{}u\allowbreak{}f\allowbreak{}l\allowbreak{}b\allowbreak{}l\allowbreak{}h\allowbreak{}n\allowbreak{}d\allowbreak{}x\allowbreak{}m\allowbreak{}l\allowbreak{}w\allowbreak{}x\allowbreak{}g\allowbreak{}s\allowbreak{}t\allowbreak{}t\allowbreak{}t\allowbreak{}v\allowbreak{}u\allowbreak{}u\allowbreak{}r\allowbreak{}j\allowbreak{}x\allowbreak{}b\allowbreak{}h\allowbreak{}c\allowbreak{}m\allowbreak{}v\allowbreak{}c\allowbreak{}w\allowbreak{}\charspacetok{}\allowbreak{}n\allowbreak{}v\allowbreak{}v\allowbreak{}r\allowbreak{}v\allowbreak{}p\allowbreak{}t\allowbreak{}p\allowbreak{}n\allowbreak{}f\allowbreak{}x\allowbreak{}b\allowbreak{}r\allowbreak{}f\allowbreak{}z\allowbreak{}m\allowbreak{}n\allowbreak{}p\allowbreak{}r\allowbreak{}b\allowbreak{}x\allowbreak{}a\allowbreak{}m\allowbreak{}t\allowbreak{}m\allowbreak{}v\allowbreak{}a\allowbreak{}n\allowbreak{}d\allowbreak{}l\allowbreak{}i\allowbreak{}l\allowbreak{}v\allowbreak{}\charspacetok{}\allowbreak{}h\allowbreak{}b\allowbreak{}i\allowbreak{}a\allowbreak{}v\allowbreak{}p\allowbreak{}c\allowbreak{}n\allowbreak{}x\allowbreak{}t\allowbreak{}k\allowbreak{}w\allowbreak{}r\allowbreak{}v\allowbreak{}n\allowbreak{}a\allowbreak{}k\allowbreak{}j\allowbreak{}k\allowbreak{}q\allowbreak{}y\allowbreak{}b\allowbreak{}v\allowbreak{}j\allowbreak{}m\allowbreak{}x\allowbreak{}m\allowbreak{}s\allowbreak{}h\allowbreak{}v\allowbreak{}u\allowbreak{}t\allowbreak{}v\allowbreak{}l\allowbreak{}e\allowbreak{}s\allowbreak{}q\allowbreak{}g\allowbreak{}y\allowbreak{}a\allowbreak{}y\allowbreak{}z\allowbreak{}d\allowbreak{}j\allowbreak{}f\allowbreak{}y\allowbreak{}e\allowbreak{}q\allowbreak{}y\allowbreak{}g\allowbreak{}l\allowbreak{}u\allowbreak{}\charspacetok{}\allowbreak{}e\allowbreak{}w\allowbreak{}p\allowbreak{}\\
\end{tabular}
}
\caption{Generations over multiple denoising steps from uniform D3PM model trained on text8 with $T=1000$. `\charspacetok{}' is the space character.}
\end{figure}

\begin{figure}
\centering
\framebox{%
\scriptsize%
}
\caption{Generations over multiple denoising steps from absorbing-state D3PM model trained on text8 with $T=1000$. `\charspacetok{}' is the space character and `\charmasktok{}' the absorbing (mask) state.}
\end{figure}

\begin{figure}
\centering
\framebox{%
\scriptsize\begin{tabular}{rp{0.95\linewidth}}
%%%%%%%%%%%%%%%%%%%%%%%%%%%%%%%%%%%%%%%%%%%%%%%%%%
\textbf{999} &\strut{}h\allowbreak{}n\allowbreak{}h\allowbreak{}f\allowbreak{}x\allowbreak{}e\allowbreak{}\charspacetok{}\allowbreak{}\charspacetok{}\allowbreak{}r\allowbreak{}c\allowbreak{}n\allowbreak{}u\allowbreak{}w\allowbreak{}h\allowbreak{}i\allowbreak{}d\allowbreak{}o\allowbreak{}r\allowbreak{}\charspacetok{}\allowbreak{}z\allowbreak{}p\allowbreak{}l\allowbreak{}u\allowbreak{}p\allowbreak{}l\allowbreak{}p\allowbreak{}a\allowbreak{}r\allowbreak{}y\allowbreak{}m\allowbreak{}d\allowbreak{}n\allowbreak{}\charspacetok{}\allowbreak{}c\allowbreak{}h\allowbreak{}q\allowbreak{}p\allowbreak{}v\allowbreak{}i\allowbreak{}j\allowbreak{}x\allowbreak{}e\allowbreak{}y\allowbreak{}w\allowbreak{}x\allowbreak{}l\allowbreak{}n\allowbreak{}k\allowbreak{}\charspacetok{}\allowbreak{}\charspacetok{}\allowbreak{}u\allowbreak{}w\allowbreak{}\charspacetok{}\allowbreak{}t\allowbreak{}g\allowbreak{}j\allowbreak{}q\allowbreak{}c\allowbreak{}\charspacetok{}\allowbreak{}q\allowbreak{}\charspacetok{}\allowbreak{}m\allowbreak{}i\allowbreak{}x\allowbreak{}p\allowbreak{}w\allowbreak{}m\allowbreak{}j\allowbreak{}n\allowbreak{}m\allowbreak{}n\allowbreak{}c\allowbreak{}o\allowbreak{}n\allowbreak{}f\allowbreak{}m\allowbreak{}d\allowbreak{}d\allowbreak{}l\allowbreak{}g\allowbreak{}z\allowbreak{}q\allowbreak{}c\allowbreak{}z\allowbreak{}c\allowbreak{}w\allowbreak{}l\allowbreak{}z\allowbreak{}n\allowbreak{}v\allowbreak{}w\allowbreak{}r\allowbreak{}s\allowbreak{}y\allowbreak{}j\allowbreak{}f\allowbreak{}\charspacetok{}\allowbreak{}b\allowbreak{}g\allowbreak{}e\allowbreak{}t\allowbreak{}a\allowbreak{}d\allowbreak{}i\allowbreak{}e\allowbreak{}a\allowbreak{}g\allowbreak{}j\allowbreak{}m\allowbreak{}t\allowbreak{}p\allowbreak{}a\allowbreak{}t\allowbreak{}l\allowbreak{}j\allowbreak{}w\allowbreak{}\charspacetok{}\allowbreak{}j\allowbreak{}p\allowbreak{}i\allowbreak{}i\allowbreak{}t\allowbreak{}i\allowbreak{}w\allowbreak{}x\allowbreak{}\charspacetok{}\allowbreak{}g\allowbreak{}f\allowbreak{}j\allowbreak{}i\allowbreak{}\charspacetok{}\allowbreak{}v\allowbreak{}c\allowbreak{}d\allowbreak{}s\allowbreak{}l\allowbreak{}k\allowbreak{}h\allowbreak{}r\allowbreak{}a\allowbreak{}h\allowbreak{}v\allowbreak{}c\allowbreak{}o\allowbreak{}k\allowbreak{}w\allowbreak{}t\allowbreak{}\charspacetok{}\allowbreak{}i\allowbreak{}y\allowbreak{}s\allowbreak{}r\allowbreak{}i\allowbreak{}z\allowbreak{}j\allowbreak{}a\allowbreak{}r\allowbreak{}r\allowbreak{}m\allowbreak{}q\allowbreak{}u\allowbreak{}h\allowbreak{}y\allowbreak{}s\allowbreak{}\charspacetok{}\allowbreak{}p\allowbreak{}d\allowbreak{}\charspacetok{}\allowbreak{}y\allowbreak{}w\allowbreak{}e\allowbreak{}i\allowbreak{}\charspacetok{}\allowbreak{}x\allowbreak{}o\allowbreak{}i\allowbreak{}j\allowbreak{}g\allowbreak{}e\allowbreak{}e\allowbreak{}g\allowbreak{}f\allowbreak{}z\allowbreak{}w\allowbreak{}l\allowbreak{}z\allowbreak{}y\allowbreak{}t\allowbreak{}r\allowbreak{}f\allowbreak{}h\allowbreak{}d\allowbreak{}\charspacetok{}\allowbreak{}p\allowbreak{}w\allowbreak{}\charspacetok{}\allowbreak{}t\allowbreak{}h\allowbreak{}s\allowbreak{}q\allowbreak{}p\allowbreak{}r\allowbreak{}l\allowbreak{}e\allowbreak{}z\allowbreak{}l\allowbreak{}h\allowbreak{}q\allowbreak{}j\allowbreak{}i\allowbreak{}s\allowbreak{}k\allowbreak{}f\allowbreak{}g\allowbreak{}p\allowbreak{}y\allowbreak{}n\allowbreak{}\charspacetok{}\allowbreak{}x\allowbreak{}r\allowbreak{}s\allowbreak{}h\allowbreak{}\charspacetok{}\allowbreak{}q\allowbreak{}\charspacetok{}\allowbreak{}f\allowbreak{}n\allowbreak{}r\allowbreak{}n\allowbreak{}o\allowbreak{}k\allowbreak{}k\allowbreak{}j\allowbreak{}q\allowbreak{}l\allowbreak{}f\allowbreak{}c\allowbreak{}c\allowbreak{}y\allowbreak{}q\allowbreak{}u\allowbreak{}a\allowbreak{}e\allowbreak{}y\allowbreak{}o\allowbreak{}r\allowbreak{}g\allowbreak{}l\allowbreak{}g\allowbreak{}a\allowbreak{}b\allowbreak{}y\allowbreak{}x\allowbreak{}o\allowbreak{}o\allowbreak{}x\allowbreak{}\\
\end{tabular}
}
\caption{Generations over multiple denoising steps from character-level nearest-neighbor D3PM model trained on text8 with $T=1000$. `\charspacetok{}' is the space character.}
\end{figure}

\end{document}